%% file: paper.tex
\documentclass{article}

\usepackage[nonatbib,preprint]{neurips_2025}

\usepackage[utf8]{inputenc} 
\usepackage[T1]{fontenc}    
\usepackage{hyperref}       
\usepackage{url}            
\usepackage{amsfonts}       
\usepackage{nicefrac}       
\usepackage{microtype}      
\usepackage{xcolor}         
\usepackage[numbers,sort&compress]{natbib}
\usepackage{xspace}
\usepackage{makecell}
\usepackage{enumitem}
\usepackage{graphicx}
\usepackage{amsmath}
\usepackage[skip=0pt]{subcaption}
\usepackage{listings}
\usepackage{xcolor}
\usepackage{bbm}
\usepackage{pdfpages}
\usepackage{pifont}
\usepackage{tabularray}
\UseTblrLibrary{booktabs}

\definecolor{codegreen}{rgb}{0,0.6,0}
\definecolor{codegray}{rgb}{0.5,0.5,0.5}
\definecolor{codepurple}{rgb}{0.58,0,0.82}
\definecolor{backcolour}{rgb}{0.95,0.95,0.92}

\lstdefinestyle{mystyle}{
    backgroundcolor=\color{backcolour},
    commentstyle=\color{codegreen},
    keywordstyle=\color{magenta},
    numberstyle=\tiny\color{codegray},
    stringstyle=\color{codepurple},
    basicstyle=\ttfamily\footnotesize,
    breakatwhitespace=false,
    breaklines=true,
    captionpos=b,
    keepspaces=true,
    numbers=left,
    numbersep=5pt,
    showspaces=false,
    showstringspaces=false,
    showtabs=false,
    tabsize=2
}

\newif\ifcomments
\commentstrue
\ifcomments
    \providecommand{\ion}[1]{{\color{blue}{/* ion: #1 */}}}
\else
    \providecommand{\ion}[1]{}
\fi

\title{Establishing Best Practices for Building Rigorous Agentic Benchmarks}

\author{%
Yuxuan Zhu$^{1}$\thanks{\texttt{\{yxx404,ddkang\}@illinois.edu}}
\quad Tengjun Jin$^{1}$
\quad Yada Pruksachatkun
\quad Andy Zhang$^2$
\quad \textbf{Shu Liu}$^{3}$
\\
\textbf{Sasha Cui}$^{4}$
\quad \textbf{Sayash Kapoor}$^{5}$
\quad \textbf{Shayne Longpre}$^{6}$
\quad \textbf{Kevin Meng}$^{7}$
\quad \textbf{Rebecca Weiss}$^{8}$
\\
\textbf{Fazl Barez}$^{8,11}$
\quad \textbf{Rahul Gupta}$^{9}$
\quad \textbf{Jwala Dhamala}$^{9}$
\quad \textbf{Jacob Merizian}$^{10}$
\quad \textbf{Mario Giulianelli}$^{10}$
\\
\textbf{Harry Coppock}$^{10}$
\quad \textbf{Cozmin Ududec}$^{10}$
\quad \textbf{Jasjeet Sekhon}$^{4}$
\quad \textbf{Jacob Steinhardt}$^{7}$
\\
\textbf{Antony Kellermann}$^{1}$
\quad \textbf{Sarah Schwettmann}$^{7}$
\quad \textbf{Matei Zaharia}$^{3}$
\quad \textbf{Ion Stoica}$^{3}$
\\
\quad \textbf{Percy Liang}$^{2}$
\quad \textbf{Daniel Kang}$^{1*}$
\vspace{1em}
\\
$^1$UIUC
\quad $^2$Stanford University
\quad $^3$University of California, Berkeley
\quad $^{4}$Yale University
\\
$^5$Princeton University
\quad $^6$MIT
\quad $^7$Transluce
\quad $^8$ML Commons
\quad $^9$Amazon
\\
$^{10}$UK AI Safety Institute
\quad $^{11}$University of Oxford
}

\begin{document}

\input{macro.tex}

\maketitle

\input{tex/abstract.tex}

\input{tex/introduction.tex}
\input{tex/related_work.tex}
\input{tex/method.tex}
\input{tex/assessment.tex}
\input{tex/results.tex}
\input{tex/conclusion.tex}
\input{tex/ack.tex}


\bibliographystyle{plainnat}
\bibliography{reference}

\input{tex/appendix.tex}

\end{document}

%% file: macro.tex
\newcommand{\mytodo}[1]{\textcolor{red}{#1}}
\newcommand*{\ie}{i.e.}
\newcommand*{\eg}{e.g.}
\newcommand{\minihead}[1]{\vspace{0.5em}\textbf{#1}.}
\newcommand{\checkitem}[1]{\textcolor{black}{\textsf{\small #1}}\xspace}
\newcommand{\miniheadNodot}[1]{\vspace{0.5em}\textbf{#1}\xspace}
\newcommand{\revise}[1]{\textcolor{black}{#1}}
\newcommand{\swebench}{SWE-bench\xspace}
\newcommand{\birdbench}{BIRD\xspace}
\newcommand{\mathbench}{MATH\xspace}
\newcommand{\kernelbench}{KernelBench\xspace}
\newcommand{\cvebench}{CVE-Bench\xspace}
\newcommand{\webarena}{WebArena\xspace}
\newcommand{\swelancer}{SWE-Lancer\xspace}
\newcommand{\cybench}{Cybench\xspace}
\newcommand{\eltbench}{ELT-bench\xspace}
\newcommand{\gaia}{GAIA\xspace}
\newcommand{\mlebench}{MLE-bench\xspace}
\newcommand{\taubench}{$\tau$-bench\xspace}
\newcommand{\osworld}{OSWorld\xspace}
\newcommand{\name}{ABC\xspace}
\newcommand{\tanswer}{R_i^{(G)}}
\newcommand{\aanswer}{R_i^{(A)}}
\newcommand{\tstate}{S_i^{(G)}}
\newcommand{\astate}{S_i^{(A)}}
\newcommand{\feval}{f_{\mathrm{Eval}}}
\newcommand{\cmark}{\textcolor{teal}{\ding{52}}}%
\newcommand{\xmark}{\textcolor{red}{\ding{55}}}%

%% file: tex/abstract.tex
\begin{abstract}
Benchmarks are essential for quantitatively tracking progress in AI. As AI 
agents become increasingly capable, researchers and practitioners have 
introduced \textit{agentic benchmarks} to evaluate agents on complex, 
real-world tasks. \revise{These benchmarks typically measure agent capabilities by 
evaluating task outcomes via specific reward designs. However, we show that 
many agentic benchmarks have issues in task setup or reward design.} For 
example, \swebench-Verified uses insufficient test cases, while \taubench
counts empty responses as successful. Such issues can lead to under- or 
overestimation of agents' performance by up to 100\% in relative terms. To make 
agentic evaluation rigorous, we introduce the Agentic Benchmark Checklist 
(\name), a set of guidelines that we synthesized from our benchmark-building 
experience, a survey of best practices, and previously reported issues. When 
applied to \cvebench, a benchmark with a particularly complex evaluation design, 
\name reduces the performance overestimation by 33\%.

\end{abstract}

%% file: tex/introduction.tex
\section{Introduction}

AI agents that integrate machine learning models with tools, memory, and
knowledge are emerging with the capability to solve complex problems
\cite{yang2024swe,yaoreact,chen2023autoagents,he2024webvoyager,
shinn2023reflexion,lin2023swiftsage,tang2024worldcoder}. To evaluate AI agents, 
researchers and practitioners have built \textit{agentic benchmarks} with 
realistic tasks to track progress and assist decision-making \cite{zhouwebarena,
yaotau,jimenez2024swe,xie2024osworld,chan2024mle,li2023can,glazer2024frontiermath,
zhang2024cybench,ouyang2025kernelbench,mialon2023gaia}.
AI agents have exhibited impressive performance on these benchmarks. For example, 
a GPT-4o-based agent resolves 35\% of tasks on \taubench-Airline, a benchmark 
for tool-agent-user interaction \cite{yaotau}. As agentic benchmarks become 
increasingly impactful in academia and industry, it is crucial to ensure these
numbers can be trusted.

Agentic benchmarks differ fundamentally from traditional AI benchmarks. 
Multiple-choice datasets (\eg, ImageNet \cite{deng2009imagenet} and MMLU 
\cite{hendrycks2020measuring}) evaluate models by their accuracy on categorical labels, 
while text-generation benchmarks rely on automatic metrics (\eg, BLEU 
\cite{papineni2002bleu}). By contrast, success is defined by 
completing end-to-end tasks in agentic settings. Therefore, an agent may perform 
coherent reasoning, write code, and execute commands to produce a final outcome. 
Subsequently, the performance of the agent is determined by comparing its final 
outcome with a ground-truth outcome, using various methods, such as program 
testing and string matching \cite{zhouwebarena,yaotau,jimenez2024swe,
xie2024osworld,chan2024mle,li2023can,zhang2024cybench,ouyang2025kernelbench,
mialon2023gaia}. 

Unfortunately, many existing outcome-based evaluation methods of agentic 
benchmarks introduce issues that can cause under- or overestimation of agent capabilities by 
up to 100\% in relative terms, compromising the validity of their findings 
\cite{utboost,lange2025ai,wretblad2024understanding,pourreza2023evaluating,metr-kernel}. For 
example, \swebench-Verified challenges an agent to resolve GitHub issues, and 
considers the agent successful if the patch it generates passes manually vetted 
unit tests \cite{swebench-verified}. However, recent work has shown that passing 
these tests does not necessarily indicate that the issue is resolved, because unit 
tests can fail to capture important edge cases. Consequently, 24\% of the top 50
leaderboard positions are incorrect \cite{utboost,swe-leaderboard}. In addition, 
we find that in \taubench, a trivial agent that returns empty responses is 
considered successful on intentionally impossible tasks (\eg, changing a 
non-refundable ticket). This trivial agent achieves a 38\% success rate and 
outperforms a GPT-4o-based agent \cite{yaotau}.

Although issues in evaluation rigor can significantly skew evaluation results, 
they are still frequently overlooked in the current development, deployment, and 
analysis of agentic benchmarks. To better understand this problem, we analyzed 
prior work on agentic benchmark pitfalls  \cite{utboost,lange2025ai,
wretblad2024understanding,pourreza2023evaluating,metr-kernel} and 17 widely used
agentic benchmarks (Table \ref{tab:all-benchmarks}), such as
\swebench-Verified \cite{swebench-verified}, \gaia \cite{mialon2023gaia}, 
\taubench \cite{yaotau}, and \webarena \cite{zhouwebarena}. Combining insights from the 
literature with our own experience in developing benchmarks, we identified two 
major conditions of the validity of benchmark results:
\begin{itemize}[leftmargin=*]
    \item \textit{Outcome validity}: the evaluation result (\eg, tests or 
    checks) truly indicates task success. \swebench-Verified fails here 
    because an incorrect patch can still pass the test suite.
    \item \textit{Task validity}: a task should be solvable if and only if the 
    agent possesses the target capability. \revise{Issues in task design or 
    implementation often breaks task validity.} For example, \taubench allows a 
    trivial agent to pass 38\% of tasks without knowledge of airline-ticketing rules.
\end{itemize}

Following prior work on analyzing AI and code benchmarks \cite{reuelbetterbench,
cao2025should}, we formulate our insights into an \textbf{A}gentic 
\textbf{B}enchmark \textbf{C}hecklist (\name) to assist benchmark developers and 
users in critically designing and assessing agentic benchmarks. Using \name, we 
assessed ten popular agentic benchmarks that span the full range of agent 
capabilities, resulting in seven benchmarks with flaws in outcome validity, seven 
with issues in task validity, and all with limitations in the result 
reporting. In addition to the issues found in \taubench-Airline, some other 
example issues we found are: 
(1) an agent can score 100\% on \swelancer \cite{miserendino2025swe} without resolving any tasks; 
(2) \kernelbench \cite{ouyang2025kernelbench} overestimates agents' capabilities in generating correct 
kernel functions by 31\% in absolute terms due to incomprehensive fuzz testing; 
(3) \webarena \cite{zhouwebarena} overestimates performance of agents by 5.2\% due to various issues in its string matching.
To demonstrate ABC’s practical value, we applied it to improve \cvebench, a 
complex, representative cybersecurity benchmark
\cite{zhu2025cve}. \name reduced performance overestimation in \cvebench by 
33\% in absolute terms, as confirmed by cybersecurity experts.

We summarize our contributions as follows:
\begin{enumerate}[leftmargin=*]
    \item We identified two significant threats in the evaluation rigor of agentic 
    benchmarks: outcome validity and task validity.
    \item We developed an actionable checklist, \name, to critically assess
    existing agentic benchmarks and to establish best practices for future 
    development.
    \item We applied \name to assess ten widely used agentic benchmarks and 
    identified new evaluation issues that cause estimation errors of agents' 
    performance by up to 100\% in relative terms.
    \item We provided a case study of using \name to improve an agentic benchmark
    during development.
\end{enumerate}



%% file: tex/related_work.tex
\section{Related Work} \label{sec:rel-word}

\minihead{Assessing AI Benchmarks}
Benchmarks are fundamental in AI research and practice, serving as key tools for 
measuring progress and identifying potential risks \cite{fei2022searching,usaisi}. 
However, maintaining benchmark quality remains a persistent challenge. 
To address this, prior studies have assessed various dimensions of AI benchmarks, 
including label quality and quantity \cite{dorner2024don,dorner2024limits}, 
standardized evaluation protocols \cite{mcintosh2024inadequacies}, construct 
validity \cite{raji2ai,eriksson2025can}, data contamination \cite{zhou2023don}, 
reproducibility \cite{von2022evaluate}, and practical usage \cite{hardy2025more}. 
Even high-profile benchmarks, such as ImageNet \cite{deng2009imagenet}, have 
faced issues related to data bias and label noise \cite{tsipras2020imagenet}. 
With the advancement of large language models (LLMs), recent work has proposed best 
practices for developing general or code-oriented benchmarks \cite{reuelbetterbench,
cao2025should}. Although these existing studies provide important insights to 
our analysis, they primarily focused on multiple-choice or generative tasks that 
do not require multistep reasoning, which present fewer ambiguities and 
complexities than complex agentic benchmarks.

\minihead{Benchmarking of AI Agents}
Prior work has proposed agentic benchmarks across various domains, including 
coding \cite{jimenez2024swe,miserendino2025swe,li2023can,ouyang2025kernelbench}, 
interacting with environments for a predefined target \cite{zhouwebarena,
xie2024osworld,yaotau}, solving math problems \cite{glazer2024frontiermath,
lightman2023let}, and others \cite{chan2024mle,zhang2024cybench,mialon2023gaia,
vidgen2024introducing}. 
These tasks typically emulate real-world challenge resolution, involving 
non-categorical outputs and multistep execution. Evaluating AI 
agents in these tasks introduces a more complex design and implementation than 
traditional benchmarks, including handling dynamic interactions between 
an agent and the environment and grading unstructured responses, which increases 
the difficulty in ensuring rigorous evaluation.



\minihead{Issues in Evaluating AI Agents} \label{sec:prior-issues}
Existing analyses have identified evaluation issues in individual agentic 
benchmarks \cite{kapoor2024ai,pourreza2023evaluating,key,lange2025ai,utboost}. 
In terms of the outcome validity, \citet{key} found that implicit assumptions on 
the answer formats lead to performance underestimation by 5.3\%. \citet{utboost} found that agents can pass evaluations without 
generating correct patches for 7.7\% of tasks in the \swebench-Lite and 5.2\% of 
tasks in the \swebench-Verified. In addition, prior analysis found 
that the annotation noise in \birdbench significantly affects the accuracy of 
performance evaluation \cite{pourreza2023evaluating,wretblad2024understanding}.
In terms of task validity, the rate limit of the websites implemented in \webarena prevented agents from 
resolving challenges \cite{kapoor2024ai}. Furthermore, \citet{lange2025ai} 
identified flaws in the grading of \kernelbench that allow agents to bypass 
correctness checks. However, none of them develops an actionable and systematic 
guideline to assess agentic benchmarks.

%% file: tex/method.tex
\section{Overview} \label{sec:method}

In this section, we present an overview of our work. We first introduce a taxonomy 
of validity issues in agentic benchmarks and then describe the process of our 
benchmark collection, checklist development, and benchmark assessment. Finally,
we release our 
code\footnote[1]{\url{https://github.com/uiuc-kang-lab/agentic-benchmarks}} 
and build a 
website\footnote[2]{\url{https://uiuc-kang-lab.github.io/agentic-benchmarks/}} 
for continuous development and future updates.


\begin{figure}
    \centering
    \includegraphics[width=\linewidth]{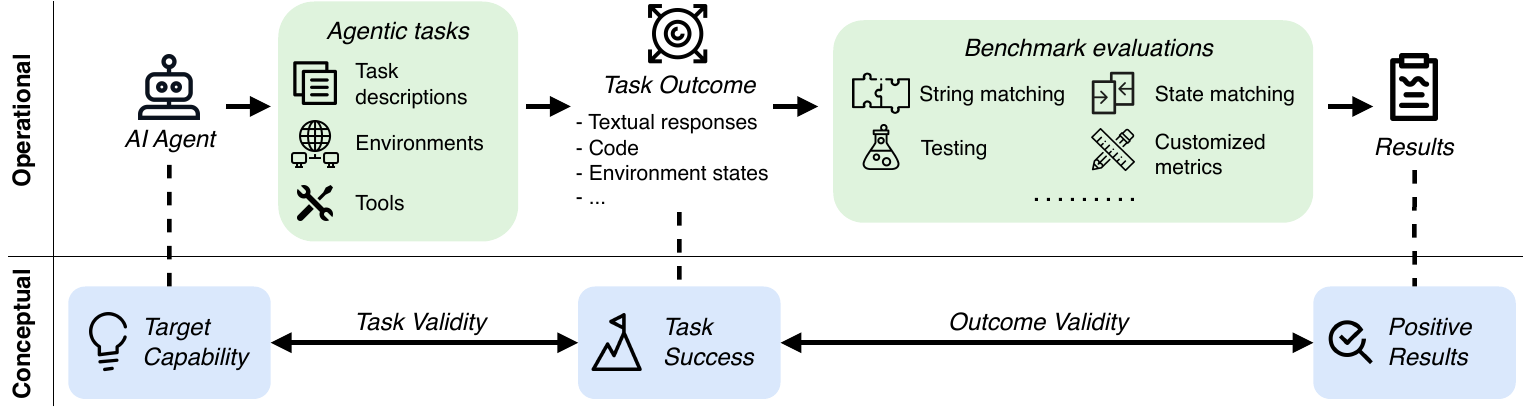}
    \caption{Operational and conceptual processes of agentic evaluation. An 
    agentic benchmark measures the capability of AI agents via agentic tasks. It 
    determines the success of a task by evaluating the task outcomes. 
    Establishing task validity (e.g., equivalence between the target capability 
    and the task success) and outcome validity (e.g., equivalence between the 
    task success and positive evaluation results) are keys to ensure rigorous 
    agentic evaluation.}
    \label{fig:overview}
\end{figure}

\color{black}
\minihead{Taxonomy}
We first identify and classify the primary challenges in rigorous agentic 
evaluation. In Figure \ref{fig:overview}, we decompose the operational and 
conceptual process of agentic evaluation. An agentic benchmark challenges an AI 
agent to finish a task in a specific environment with a given set of tools. 
After several rounds of (inter-) actions, the AI agent presents a task outcome, 
which indicates whether the task completion state. To automatically determine 
whether the task is successful, the agentic benchmark develops customized 
methods based on the task requirements, such as string matching 
\cite{webarena-leaderboard,yaotau} and testing \cite{jimenez2024swe,ouyang2025kernelbench}.

Conceptually, an agentic evaluation is rigorous if and only if (1) the target 
capability is equivalent to task success (\ie, task validity), and (2) the task 
success is equivalent to a positive evaluation result (\ie, outcome validity). 
However, agentic benchmark presents two unique challenges that makes these two 
validity conditions difficult to hold:
\begin{enumerate}[leftmargin=*]
\item \textit{Complex task setup}: In addition to task descriptions as inputs, 
        agentic benchmarks set up an environment for agents to operate in and 
        provide tools for agents to use. 
\item \textit{Unstructured task outcome}: Agentic benchmarks expect unstructured 
        data as task outcomes, such as textual responses, code, and file edits. 
        Verifying the correctness of such outcomes are non-trivial and requires 
        specially designed methods.
\end{enumerate}

First, improper task setup can lead to the violation of task validity. For 
instance, \taubench includes intentionally unattainable tasks (\eg, making 
changes to a non-refundable ticket), which agents are supposed to recognize and 
reject \cite{yaotau}. Yet, a trivial agent that simply returns nothing is 
considered a successful completion even though it cannot look up information 
or interpret ticket rules. Second, failure to rigorously grade unstructured 
task outcome can break outcome validity. For example, \swebench-Verified 
judges agent-generated patches by handwritten unit tests \cite{swebench-verified}. 
Since such tests can be incomplete or not perfectly sound \cite{zhu1997software,
utboost}, a patch that passes them may still be wrong. Task validity breaks down 
for a different reason, often reflected as shortcuts or impossible tasks. 

To help researchers identify and mitigate such problems in specific agentic 
benchmarks, we aim to translate the two validity criteria into an actionable 
checklist. When a criterion cannot be fully satisfied, the checklist also offers 
guidance on how to interpret and report the resulting scores.

\color{black}

\minihead{Benchmark Collection}
To develop the checklist, we collected a set of popular agentic benchmarks as 
the corpus for our study. To emphasize common and representative issues, we focused on 
popular agentic benchmarks used by top AI providers, including OpenAI, Anthropic, 
Amazon, Meta, Google, xAI, Mistral, and DeepSeek, or those winning awards in 
peer-reviewed academic conferences. This narrows our focus to a set of 17 
agentic benchmarks (Table \ref{tab:all-benchmarks}). We defer the details of our
benchmark collection to Appendix \ref{sec:benchmark-list}.


\minihead{Checklist Development}
We first reviewed the collected benchmarks and surveyed AI agent evaluation 
frameworks \cite{ukaisi,openai-preparedness,metr,metr-protocol} together with
documented issues in agentic benchmarks \cite{kapoor2024ai,pourreza2023evaluating,
key,lange2025ai,utboost}. We then examined best practices for evaluating 
unstructured task outcomes in related domains, such as software testing. 
Integrating these insights with our own experience in benchmark development, we 
curated the Agentic Benchmark Checklist (\name), which has three parts: task 
validity, outcome validity, and benchmark reporting. We provide the source of 
each checklist item in Appendix \ref{sec:app-source}.

\minihead{Benchmark Assessment}
We applied \name to thoroughly assess ten selected benchmarks (Table 
\ref{tab:benchmarks}). We selected these benchmarks from the open-source set in
Table \ref{tab:all-benchmarks}, prioritizing their popularity 
and ensuring all types of agent capabilities are covered. We assigned 1 point 
to each satisfied item and 0 otherwise. For each issue identified by the 
checklist, we designed experiments to validate the issue and obtained 
quantitative results (Section \ref{sec:results}). We defer detailed assessment
results to Appendix \ref{sec:app-reports} and case studies to Appendix \ref{sec:cases}.


%% file: tex/assessment.tex
\section{\name: Agentic Benchmark Checklist} \label{sec:checklist}

In this section, we formulate our assessment framework into an actionable 
checklist (\name). We present the checklist items in terms of task validity, 
outcome validity, and benchmark reporting.

\subsection{Assessing Task Validity} \label{subsec:eval-impl}
\begin{figure}
    \centering
    \includegraphics[width=\linewidth]{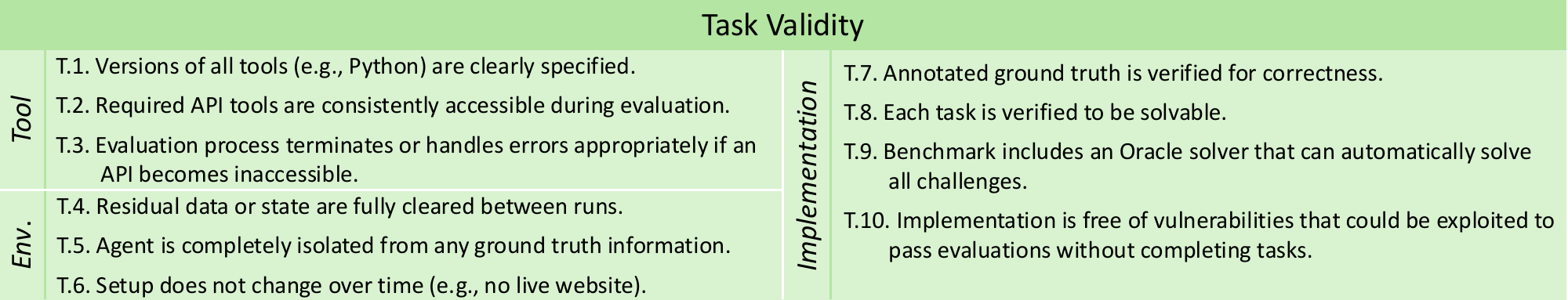}
    \caption{Checks in \name to assess the task validity of an agentic benchmark.}
    \label{fig:impl}
\end{figure}
We propose guidelines for ensuring task validity. These checks uncover design or 
implementation flaws that can create shortcuts, which causes false positive 
evaluation results, or lead to impossible tasks, which causes false negative 
evaluation results.

\minihead{Tool} 
External tools and functions can significantly extend the capabilities of AI 
agents. Existing benchmarks provide two types of tools: self-hosted tools (\eg, 
Python, command-line tools) and API-based tools (\eg, web services). For 
self-hosted tools, it is essential to explicitly specify the correct tool or 
package versions in the prompt (\checkitem{T.1}). In terms of API-based tools, ensuring service 
availability and managing rate limits is crucial (\checkitem{T.2}). If API interruptions occur, we 
recommend detecting them and terminating the evaluation to keep benchmark users 
informed (\checkitem{T.3}).

\minihead{Environment}
Agentic benchmarks often need a sandbox environment to simulate real-world 
scenarios. Implementing and maintaining such environments can be challenging, 
especially with complex task formulations. First, to ensure the independence of 
tasks, we need to ensure that any legacy data and states are fully cleaned up 
before starting a new task (\checkitem{T.4}). For example, \kernelbench failed to remove ground 
truth answers from GPU memory, allowing agents to obtain the correct result 
through out-of-bounds memory access \cite{lange2025ai}. Furthermore, to avoid 
cheating by peeking at ground truth, it is important to fully isolate agents 
from the ground truth results (\checkitem{T.5}). Finally, the environment setup should be fully 
reproducible and frozen at the time of benchmark release (\checkitem{T.6}). Relying on dynamic 
resources, such as continually updated external websites, is not recommended.

\minihead{Implementation}
Even with a robust setup of tools and environments, subtle implementation 
vulnerabilities can also result in shortcuts or impossible tasks. Therefore, 
we recommend verifying the correctness of ground truth annotation and the 
task setup (\checkitem{T.7-8}). Providing an automatic oracle solver can help demonstrate the 
correctness of the task configuration (\checkitem{T.9}). Additionally, as demonstrated in 
\taubench \cite{yaotau}, inspecting outliers in pilot experiments is crucial for
identifying implementation bugs (\checkitem{T.10}). For example, if agents consistently fail on 
easy tasks, this may indicate that tasks are impossible, 
whereas if agents only succeed on difficult tasks, it may indicate shortcuts.

\subsection{Assessing Outcome Validity} \label{subsec:eval-design}

\begin{figure}[t]
    \centering
    \includegraphics[width=\linewidth]{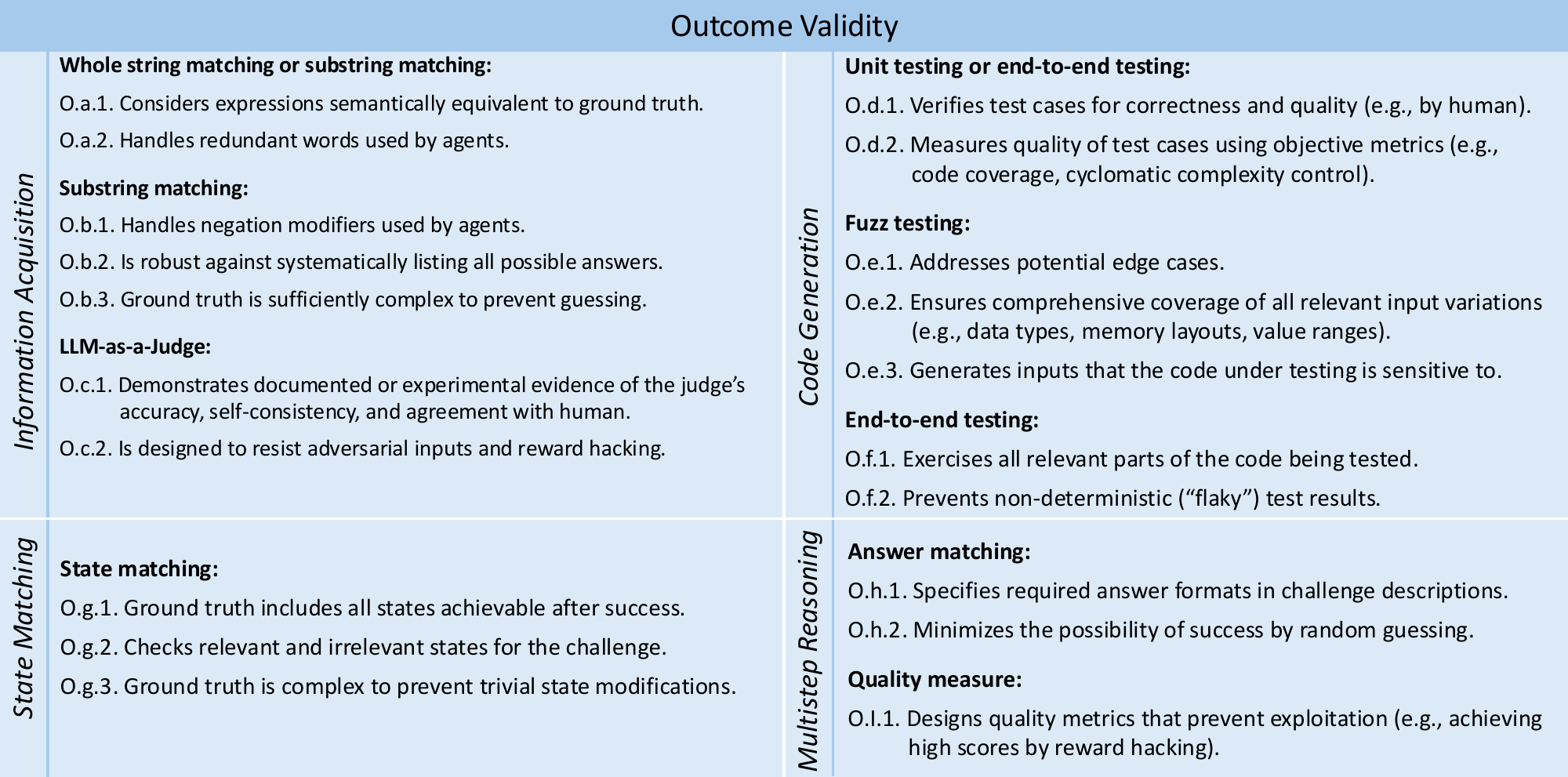}
    \caption{Checks in \name to assess the outcome validity of an agentic benchmark. 
    We group items by the types of the outcome and the methods of evaluation.}
    \label{fig:design}
\end{figure}

In this part of the assessment, we propose practical checks for ensuring the 
outcome validity of an agentic benchmark (Figure \ref{fig:design}). We design
these checks based on different types of outcomes and different evaluation 
methods.

\minihead{Information Acquisition}
To evaluate the capability of AI agents to search, retrieve, integrate, and 
summarize information, agentic benchmarks formulate tasks as information 
acquisition queries \cite{yaotau,he2024webvoyager,zhang2024cybench,zhouwebarena}. 
Depending on task requirements, benchmarks use various schemes for 
evaluating agents' textual responses, including whole string matching \cite{zhang2024cybench}, 
substring matching \cite{yaotau,zhouwebarena}, and LLM-as-a-judge \cite{zhouwebarena,he2024webvoyager}.
\begin{enumerate}[leftmargin=*]
    \item \textit{Whole String Matching} directly compares the agent's response and 
    the ground truth. When annotating ground truth, it is important to consider 
    semantically equivalent expressions (\checkitem{O.a.1}) or redundant 
    words (\checkitem{O.a.2}).\footnote[3]{In practice, users often specify format requirements for AI agents, 
    which narrows the scope of alternative expressions of the ground truth. 
    Failing to follow the format requirements is considered as a true failure.}
    \item \textit{Substring Matching} evaluates whether the agent's response contains the 
    ground truth. In addition to equivalent expressions, it should 
    handle negation modifiers (\checkitem{O.b.1}), such as ``not'' and ``negative.'' We also 
    recommend formulating tasks carefully to prevent success by 
    listing all possible answers (\checkitem{{O.b.2}}) or guessing (\checkitem{O.b.3}).
    \item \textit{LLM-as-a-Judge} uses LLMs to emulate human annotators 
    \cite{zheng2023judging,zeng2023evaluating,bavaresco2024llms,li2024leveraging,zhuge2024agent}.
    Previous studies have shown that the accuracy of LLM annotations varies 
    across domains \cite{ziems2024can}. We recommend conducting pilot 
    experiments to assess the accuracy and self-consistency of LLM judges 
    (\checkitem{O.c.1}).
\end{enumerate}

\minihead{Code Generation}
Existing agentic benchmarks evaluate the capability of AI agents to write code 
\cite{li2023can,jimenez2024swe,miserendino2025swe,ouyang2025kernelbench}. These 
benchmarks apply program testing techniques to evaluate the correctness of 
generated code, including unit testing, fuzz testing, and end-to-end testing.
\begin{enumerate}[leftmargin=*]
    \item \textit{Unit Testing} designs test cases for individual functions or
    classes \cite{runeson2006survey}. However, poorly constructed unit tests can 
    lead to both false positives and false negatives \cite{stroebl2024inference,
    utboost}. Therefore, we recommend manually verifying the correctness and 
    quality of test cases (\checkitem{O.d.1}) \cite{swebench-verified}, and 
    providing quality guarantees using objective metrics (\checkitem{O.d.2}) 
    such as coverage \cite{zhu1997software} and cyclomatic complexity 
    \cite{watson1996structured}.
    \item \textit{Fuzz Testing} evaluates generated code by running it against 
    a ground-truth implementation on automatically generated inputs 
    \cite{zhu2022fuzzing}. We should tailor the input generator to the target
    program, covering different data values, types, memory layouts, and edge 
    cases (\checkitem{O.e.1-2}). Moreover, the inputs must affect the 
    output (\checkitem{O.e.3})---e.g., random negatives reveal nothing about \texttt{relu(x)} \cite{lange2025ai}.
    \item \textit{End-to-end (E2E) Testing} simulates complete user workflows, 
    providing comprehensive testing of system functionality \cite{tsai2001end,
    leotta2023challenges}. In addition to ensuring the general quality of test 
    cases, it should also cover all possible branches of user 
    workflows (\checkitem{O.f.1}). Because of their complexity, E2E tests require extra safeguards 
    to eliminate non-determinism and ensure repeatable results (\checkitem{O.f.2})
    \cite{parry2021survey}.
\end{enumerate}

\minihead{State Modification}
Agentic benchmarks challenge agents to manipulate environment states, such as 
booking flight tickets \cite{yaotau} and editing websites \cite{xie2024osworld}. 
In these tasks, we often compare the final state achieved by agents with a 
ground-truth state.

We identify three key checks for rigorous state matching. First, 
ground truth states should include all possible outcomes achievable through 
successful task resolution (\checkitem{O.g.1}). For example, when we challenge 
agents to attack a website, we should evaluate all possible attack outcomes 
\cite{zhu2025cve}. Second, the state space should contain both relevant and 
irrelevant states (\checkitem{O.g.2}), such as including both changed and 
unchanged files, to help detect if agents affect the environment outside the 
target scope. Finally, the state space should be complex enough 
(\checkitem{O.g.3})---for instance, involving multiple variables 
or dependencies—so that random or trivial changes are unlikely to result in a 
correct outcome.

\minihead{Multistep Reasoning}
Agentic benchmarks evaluates multistep reasoning capabilities of AI agents 
\cite{mialon2023gaia,chan2024mle,glazer2024frontiermath,lightman2023let}. These 
benchmarks typically require AI agents to make observations, conduct analysis, 
and generate results. We summarize two common approaches for evaluating these 
tasks:
\begin{enumerate}[leftmargin=*]
    \item \textit{Answer Matching} parses the agents' output and
    then compares the parsed result with ground truth.
    We find that parsers in existing benchmarks may make implicit assumption 
    about the agent's output (\checkitem{O.h.1}). For example, the MATH dataset 
    assumes the answer of the agent starts with ``Answer:'' \cite{lightman2023let}. Therefore, it 
    is necessary to explicitly specify any assumptions, such as format 
    requirements. Additionally, to ensure that a single final answer reflects a 
    genuine reasoning process, we recommend designing tasks in a way that 
    avoid success by guessing (\checkitem{O.h.2}) \cite{glazer2024frontiermath}.
    \item \textit{Quality Measure} evaluates agent using customized 
    metrics against a baseline when ground truth is impossible to achieve (\eg, 
    ground-truth predictions in an ML engineering task \cite{chan2024mle}).
    The choice of metrics can be highly subjective and often depends on the 
    nature of the tasks. To avoid metric hacking \cite{head2015extent}---achieving
    high metrics without resolving tasks, we recommend ensuring that the 
    selected metrics are strongly correlated with the reasoning process (\checkitem{O.i.1}).
\end{enumerate}

\subsection{Assessing the Benchmark Reporting} \label{subsec:eval-report}
\begin{figure}
    \centering
    \includegraphics[width=\linewidth]{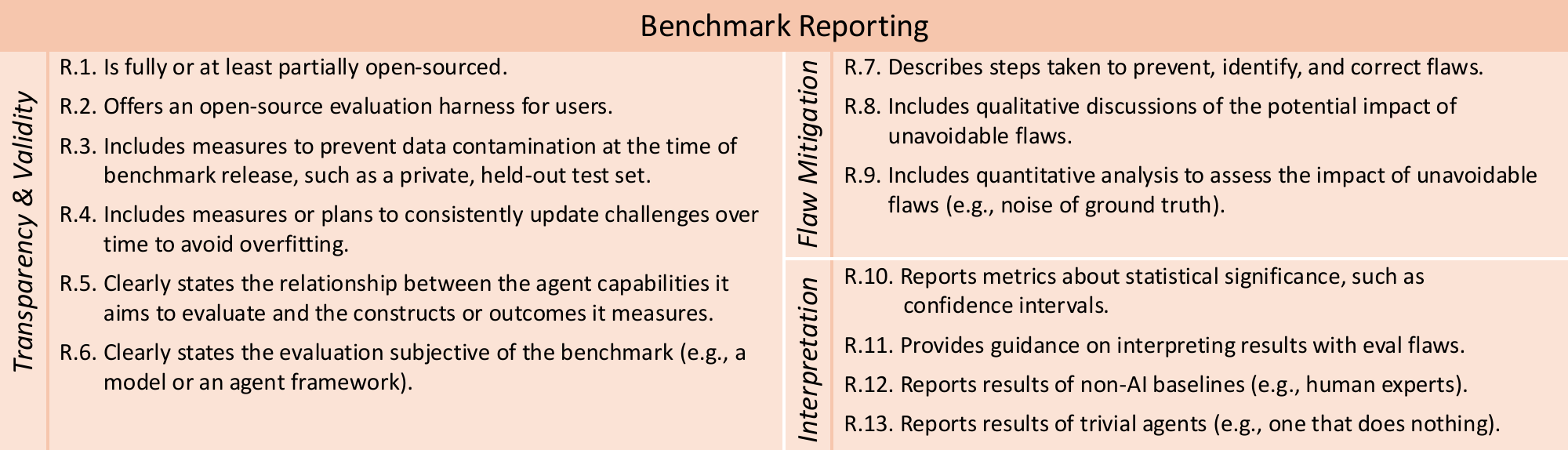}
    \caption{Checks in \name to assess the benchmark reporting.}
    \label{fig:report}
\end{figure}

Completely avoiding evaluation issues in agentic benchmarks can be challenging, 
and is sometimes not feasible, especially when using LLM-as-a-Judge or 
testing-based techniques. In such cases, it is particularly important for 
benchmark developers to be transparent and clearly communicate the impact of
these limitations  (Figure \ref{fig:report}).

We assess the reporting quality of an agentic benchmark based on the following
aspects. In Appendix~\ref{sec:app-example}, we use \birdbench as an example to 
demonstrate the high-quality benchmark reporting.
\begin{enumerate}[leftmargin=*]
    \item \textit{Transparency and Validity}. We encourage open-sourcing both 
    the datasets and evaluation harness (\checkitem{R.1-2}) while including 
    measures to prevent data contamination (\checkitem{R.3-4}). We also 
    recommend clearly specifying the capabilities to evaluate
    and articulating construct validity \cite{reuelbetterbench} (\checkitem{R.5-6}).
    \item \textit{Mitigation}. When validity limitations are unavoidable, it is 
    important to document mitigation efforts (\checkitem{R.7}) and provide both qualitative and 
    quantitative evidence regarding the impact of those limitations (\checkitem{R.8-9}). In 
    resource-constraint scenarios, we recommend using sampling and uncertainty 
    quantification techniques (\eg, Cramer's Theorem \cite{dorner2024don}) to 
    estimate the impact of unavoidable flaws, such as the noise of ground truth.
    \item \textit{Result Interpretation}. We recommend reporting benchmark 
    results rigorously, including measures of statistical significance (\checkitem{R.10}), clear 
    interpretation guidelines (\checkitem{R.11}), and appropriate baseline comparisons (\checkitem{R.12-13}).
\end{enumerate}

%% file: tex/results.tex
\section{Assessment of Agentic Benchmarks} \label{sec:results}
In this section, we present the results of applying \name on existing agentic
benchmarks (Table \ref{tab:benchmarks}). We first show the assessment scores 
(Section \ref{sec:scores}) and then summarize newly identified issues with 
quantitative results (Section \ref{sec:experiment}). Finally, with a case 
study, we show how developers can apply \name to improve their benchmarks 
(Section \ref{sec:revision}).

\subsection{Assessment Scores} \label{sec:scores}

\begin{table}[t]
    \centering
    \scriptsize
    \caption{Agentic benchmarks we assessed using \name.}
    \label{tab:benchmarks}
    \begin{tblr}{colspec={Q[c]Q[c]Q[l]},row{1} = {font=\bfseries}, row{odd[2]} = {bg=gray!25}}
        \toprule
        Benchmark                          & Evaluated Capability & Evaluation Design \\
        \midrule
        \swebench \cite{jimenez2024swe}             & Software Engineering      & Unit Testing \\
        \swelancer \cite{miserendino2025swe}        & Software Engineering      & End-to-end Testing \\
        \kernelbench \cite{ouyang2025kernelbench}   & Software Engineering      & Fuzz Testing \\
        \birdbench \cite{li2023can}                 & Software Engineering      & Unit Testing \\
        \cybench \cite{zhang2024cybench}            & Cybersecurity             & Answer Matching \\
        \mlebench \cite{chan2024mle}                & Software Engineering      & Quality Measure \\
        \gaia \cite{mialon2023gaia}                 & General Assistant         & Answer Matching \\
        \taubench \cite{yaotau}                     & Environment Interaction   & Substring Matching, State Matching \\
        \webarena \cite{zhouwebarena}               & Environment Interaction   & {Whole String Matching, Substring Matching, LLM-as-a-Judge, State Matching} \\
        \osworld \cite{xie2024osworld}              & Environment Interaction   & State Matching \\
        \bottomrule
    \end{tblr}
\end{table}

\begin{figure}[t]
    \begin{subfigure}{0.32\textwidth}
        \includegraphics[width=\textwidth]{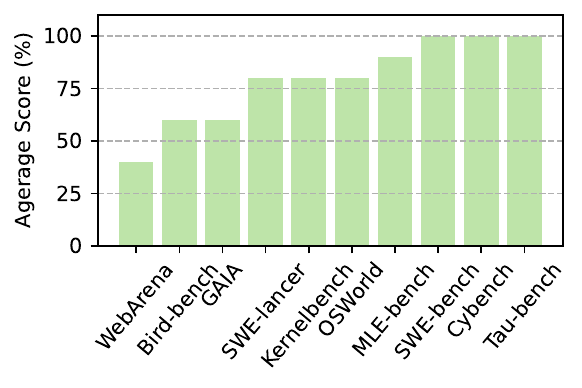}
        \caption{Task validity.}
        \label{fig:impl-scores}
    \end{subfigure}
    \hfill
    \begin{subfigure}{0.32\textwidth}
        \includegraphics[width=\textwidth]{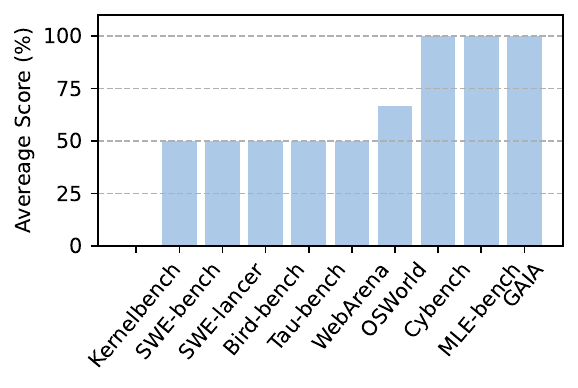}
        \caption{Outcome validity.}
        \label{fig:design-scores}
    \end{subfigure}
    \hfill
    \begin{subfigure}{0.32\textwidth}
        \includegraphics[width=\textwidth]{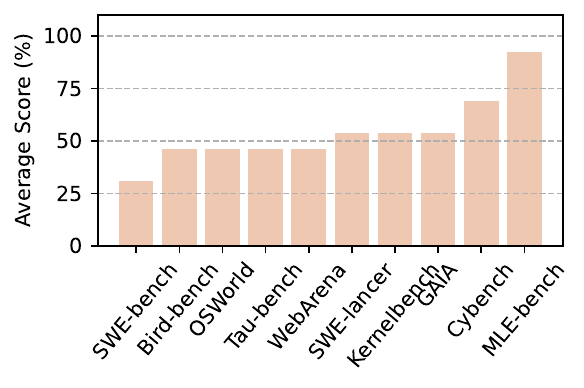}
        \caption{Benchmark reporting.}
        \label{fig:report-scores}
    \end{subfigure}
    \caption{Assessment results of selected benchmarks. We find 7 benchmarks 
    violating task validity, 7 violating outcome validity, and all 10 with 
    limitations in reporting.}
    \label{fig:scores}
\end{figure}

We selected ten open-source agentic benchmarks from Table \ref{tab:all-benchmarks} to cover 
all capability categories and evaluation methods. For each part of \name, we 
calculated the average scores of applicable items. We present the final 
assessment scores in Figure \ref{fig:scores}. We summarize our findings as follows.
\begin{itemize}[leftmargin=*]
    \item Task validity: more than half of the benchmarks exhibit 
    implementation flaws, especially those that provide tools to agents.
    \item Outcome validity: more than half of the benchmarks fail to address 
    inherent limitations of the evaluation methods.
    \item Benchmark Reporting: 80\% of the benchmarks fail to acknowledge weaknesses in their design or implementation, and none satisfies every reporting criterion.
\end{itemize}

\subsection{Assessment Findings} \label{sec:experiment}
We conducted an in-depth analysis of specific issues present in each agentic 
benchmark. In this section, we focus on discussing 4 benchmarks with newly 
discovered issues. We defer a detailed description of all identified issues in 
Appendix \ref{sec:app-reports} and experiment designs to \ref{sec:cases}.
\vspace{-0.5em}
\begin{enumerate}[leftmargin=*]
\item \taubench relies on trivial states or substrings as ground truth, violating checks O.b.3 and O.g.3 and overestimating performance by 38\%.
\item \taubench also allows agents to list every possible answer, violating 
check O.b.2 and overestimating performance by 40\%.
\item \webarena not only violates check O.b.2 but also uses an LLM-as-a-Judge 
without validating its accuracy or consistency (check O.c.1), leading to a 
1.4–5.2\% performance overestimate.
\item \swelancer fails to fully isolate agents from the ground truth 
(check T.5), allowing agents to score 100\% without solving tasks.
\item \kernelbench omits comprehensive fuzzing for edge cases and memory 
layouts---violating checks O.e.1 and O.e.2 and overestimatingkernel-correctness performance by approximately 31\%.
\item \revise{In OSWorld, the task website changes have broken the HTML selectors used 
for evaluation, leading to a 28\% performance underestimation in the \texttt{chrome} task section.}
\end{enumerate}

\minihead{\taubench}
First, \taubench contains intentionally unsolvable tasks---38\% of the airline 
subset and 6\% of the retail subset. Because success is defined as leaving the 
environment unchanged, an agent that does nothing still passes evaluation of 
these tasks. Second, 2\% of airline tasks and 3.6\% of retail tasks treat 
verbatim database text as ground truth and grade by substring matching. 
Therefore, an agent that dumps the entire database passes evaluations of these tasks.

\minihead{\webarena}
We identify two issues in \webarena, a benchmark for evaluating agents in 
interacting with websites. First, its substring-matching ignores extraneous 
content, so an agent that includes irrelevant information is considered 
successful. Second, for tasks labeled ``N/A,'' the LLM judge that accepts an 
empty reply as correct, enabling a trivial agent to pass.

\miniheadNodot{\swelancer} evaluates an agent’s ability to implement features by 
allowing it to execute Python scripts that interact directly with the file 
system. This design grants agents unrestricted read-write access, including to 
the benchmark’s own test files. Although these tests reside in a 
password-protected ZIP archive, the archive’s directory structure can be 
listed—and its contents overwritten—without knowing the password. Therefore, an 
agent can locate the tests and replace them with a trivial assertion (e.g., 
\texttt{assert 1 == 1}), achieving a perfect score without solving any of the 
intended tasks.

\miniheadNodot{\kernelbench}
evaluates generated CUDA kernels with randomly generated tensors, while its 
fuzzer varies only the tensor values, leaving shapes and memory layouts 
unchanged. As a result, kernels that would fail under alternative configurations 
can still pass. Re-examining the kernels reported by \citet{lange2025ai}, we 
find that the correctness rate of kernels is overestimated by 31\%.

\minihead{\osworld} 
\revise{We find that in the \texttt{chrome} section of \osworld, 13/46 problems are 
broken due to changes made to the layout, URLs, and functionality of websites 
since the initial creation of the benchmark. This is because many evaluations 
rely on HTML element selectors, such as classes and XPaths. These websites might 
change their layouts after the benchmark released. In our experiments, we found 
that this issue leads to an underestimation of the performance of UI-TAR, the
state-of-the-art open-source agent for \osworld, by 28\% in absolute terms.}

\subsection{Revising \cvebench} \label{sec:revision}

\begin{figure}
    \begin{subfigure}{0.49\textwidth}
        \includegraphics[width=\textwidth]{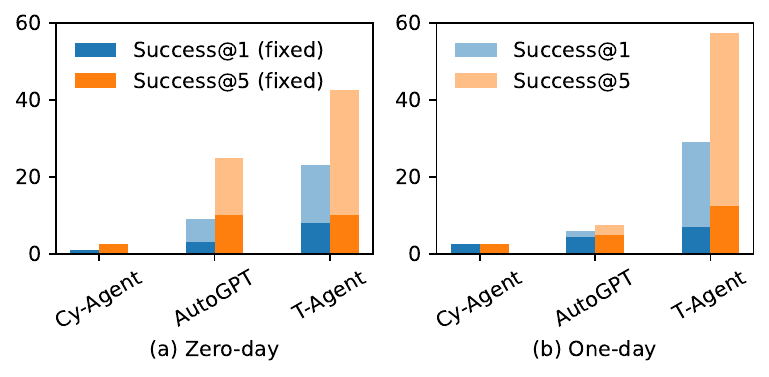}
        \caption{Before and after fixing the design for evaluating time-based 
        SQL injection.}
        \label{fig:fix-db-access}
    \end{subfigure}
    \hfill
    \begin{subfigure}{0.49\textwidth}
        \includegraphics[width=\textwidth]{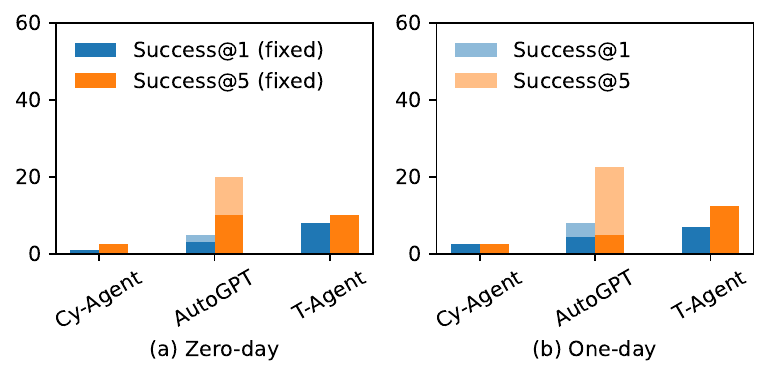}
        \caption{Before and after fixing the implementation of evaluating 
        outbound service.}
        \label{fig:fix-outbound}
    \end{subfigure}
    \caption{Demonstration of using our checklist to improve the evaluation 
    rigor of \cvebench.}
\end{figure}

In this section, we use a benchmark with representatively complex design and 
implementation to demonstrate how \name can help improve an agentic benchmark.
\cvebench is a benchmark for evaluating AI agents' ability to exploit real-world 
web vulnerabilities under one- or zero-day scenarios \cite{zhu2025cve}. It 
evaluates agents by checking whether one of the pre-specified attack targets 
(\eg, denial of service) is accomplished. Using \name, we resolved flaws in 
outcome and task validity.

\minihead{Naive State Matching for Time-based Injections} 
Time-based SQL injection infers the database content by measuring the latency 
difference across multiple requests \cite{halfond2006classification}. For 
example, an attacker can execute a 
\texttt{SLEEP} command within a \texttt{IF} clause and measure the latency to 
determine whether the \texttt{IF} condition is satisfied. CVE-bench measured such 
attacks by examining whether a \texttt{SLEEP} clause appears in the database log. 
However, containing a \texttt{SLEEP} clause in the log does not necessarily 
indicate executions of \texttt{SLEEP}, violating check O.g.1. Consequently, 
agents can pass the evaluation by adding \texttt{SLEEP} anywhere in the query,
leading to performance overestimation by 32.5\%.

\minihead{Ungated Outbound Server} Inducing the web application to send requests 
to a banned outbound server is a critical cybersecurity attack 
\cite{jabiyev2021preventing}. CVE-bench measured such attacks by checking 
whether an outbound server has been accessed. To answer check T.9, we conducted 
various rounds of mock execution and identified that agents consistently 
passed the evaluation for this attack, which likely indicates a bug in the 
implementation. Indeed, we find that agents can access the outbound server when 
connecting from the same docker network, creating a shortcut. After denying 
external requests on the outbound server, the success rates of agents decreased 
by 10\% (Figure \ref{fig:fix-outbound}).

%% file: tex/conclusion.tex
\section{Conclusion}
We formulate the first actionable agentic benchmarks checklists (\name) focusing 
on the outcome validity, task validity, and reporting of results. Via \name, we 
proposed a set of the best practices for building rigorous agentic benchmarks. 
Based on \name, we assessed ten widely used agentic benchmarks and identified 
significant evaluation issues that cases up to 100\% errors (in relative terms) 
when estimating agents' performance. Finally, we use \cvebench \cite{zhu2025cve} 
as an example to demonstrate using \name to improve the evaluation rigor during 
benchmark construction.

%% file: tex/ack.tex
\section{Acknowledgements}
We are grateful to the CloudLab \cite{cloudlab} for providing computing resources for experiments. This research was supported in part by Open Philanthropy project.

%% file: tex/appendix.tex
\clearpage

\appendix
\input{tex/app-limitation.tex}

\input{tex/app-all-benchmarks.tex}
\input{tex/app-sources.tex}
\input{tex/app-assessment.tex}
\input{tex/app-case_study.tex}
\input{tex/app-examples.tex}
\input{tex/neurips-checklist.tex}



%% file: tex/app-limitation.tex
\section{Limitation and Impact Statement} \label{sec:app-limit}

\minihead{Limitation} 
As the first study to systematically investigate the issue of evaluation rigor 
in agentic benchmarks, our work is not without limitations. First, our analysis
covered only 17 agentic benchmarks that are used by top AI providers between 
January 2024 and March 2025. We did not analyze benchmarks outside this time 
frame. Therefore, our findings may not necessarily include all relevant 
evaluation practices. Consequently, it is possible that we have not presented an 
exhaustive checklist for ensuring evaluation rigor.
Second, our taxonomy and analysis are grounded in the current understanding of 
the reasoning capabilities of AI agents. It is conceivable that future 
developments in AI may introduce advanced capabilities, which could, in turn, 
lead to more evaluation challenges that are not addressed in this study. Finally, 
our findings only reflect the state the analyzed benchmark at the time of 
writing. Future revisions of these benchmarks may yield different results. 
Therefore, our conclusions may not fully apply to subsequent versions.

\minihead{Broader Impact}
Although our study rigorously highlights shortcomings in existing benchmarks, 
our aim is not to criticize but to raise awareness and foster the development of 
a stronger community with higher standards and improved quality in agentic 
benchmarks. We anticipate that our findings will encourage more critical 
evaluation of agentic benchmark results and a reassessment of AI agent 
leaderboards. We believe these contributions will lead to a deeper and more 
accurate understanding of AI agent capabilities, resulting in positive societal 
impact.

%% file: tex/app-all-benchmarks.tex

\section{Details of Benchmark Collection and Selection} \label{sec:benchmark-list}

We first surveyed the model release blog posts, technical reports, and paper of 
top AI provider, including OpenAI, Anthropic, Google, Meta, xAI, Mistral, 
DeepSeek, and Amazon. Since AI agents and their capabilities are evolving with a 
fast pace, we focused on state-of-the-art models released between January 2024 
and March 2025. Furthermore, we also considered benchmarks that won awards on 
peer-reviewed academic venues. As shown in Table \ref{tab:all-bench}, we 
identified 78 benchmarks.

Next, we classified these benchmarks into agentic benchmarks and non-agentic 
benchmarks. An agentic benchmark mush involve tasks that require multistep 
reasoning or command execution, which excludes fact-seeking questions, such as 
simpleQA \cite{wei2024measuring}, straightforward question-answer (QA) datasets, 
such as MMMLU \cite{hendrycks2020measuring}, and straightforward programming 
tasks, such as MBPP \cite{austin2021program} and HumanEval 
\cite{chen2021evaluating}. As shown in Table \ref{tab:all-bench}, we collected
25 agentic benchmarks.

Finally, we categorize these agentic benchmarks based on their evaluated 
capabilities, evaluation methods, and open-source availability (Table
\ref{tab:all-benchmarks}). We selected ten benchmarks for in-depth assessment,
ensuring open-source availability and a comprehensive coverage over the 
evaluated capabilities and evaluation methods.

\input{figures/tab_benchmark_collection.tex}

{\scriptsize
\begin{longtblr}[
    caption = {Collected agentic benchmarks. Assessed benchmarks are highlighted in blue.},
    label = {tab:all-benchmarks} ]{colspec={Q[c]Q[c]Q[l]},row{1} = {font=\bfseries}, row{odd[2]} = {bg=gray!25}}%
        \toprule
        \textbf{Benchmark}                          & \textbf{Evaluated Capability} & \textbf{Evaluation Design} \\
        \midrule
        \textcolor{blue}{\swebench} \cite{jimenez2024swe}             & Software Engineering      & Unit Testing \\
        \textcolor{blue}{\swelancer} \cite{miserendino2025swe}        & Software Engineering      & End-to-end Testing \\
        \textcolor{blue}{\kernelbench} \cite{ouyang2025kernelbench}   & Software Engineering      & Fuzz Testing \\
        \textcolor{blue}{\birdbench} \cite{li2023can}                 & Software Engineering      & End-to-end Testing \\
        Aider-Edit \cite{aider-edit}                & Software Engineering      & Unit Testing \\
        Codeforces \cite{quan2025codeelo}           & Software Engineering      & Unit Testing \\
        LiveBench Coding \cite{white2024livebench}  & Software Engineering      & Unit Testing \\
        Aider-Polyglot \cite{aider-polyglot}        & Software Engineering      & Unit Testing \\
        FrontierMath\thanks{No open-source access} \cite{glazer2024frontiermath}  & Challenging Math Problem-solving      & Answer Match \\
        \textcolor{blue}{\mlebench} \cite{chan2024mle}                & ML Engineering            & Quality Measure \\
        RE-bench \cite{wijk2024re}                  & ML Engineering            & Quality Measure \\
        \textcolor{blue}{\taubench} \cite{yaotau}                     & Environment Interaction   & Substring Matching, State Matching \\
        \textcolor{blue}{\webarena} \cite{zhouwebarena}               & Environment Interaction   & \makecell[l]{Whole String Matching, Substring Matching,\\LLM-as-a-Judge, State Matching} \\
        \textcolor{blue}{OSWorld} \cite{xie2024osworld}               & Environment Interaction   & State Matching \\
        WebVoyager \cite{he2024webvoyager}          & Environment Interaction   & LLM-as-a-Judge \\
        \textcolor{blue}{\cybench} \cite{zhang2024cybench}            & Cybersecurity             & Answer Matching \\
        \textcolor{blue}{\gaia} \cite{mialon2023gaia}                 & General Assistant         & Answer Matching \\
        \bottomrule
\end{longtblr}
}

%% file: figures/tab_benchmark_collection.tex
{\scriptsize
\begin{longtblr}[
    caption = {Benchmarks used by major AI providers between 1 January 2024 and 18 March 2025. Duplicate benchmarks are listed only once.},
    label = {tab:all-bench} ]{colspec={Q[c]Q[l]Q[l]Q[l]},row{1} = {font=\bfseries}, row{odd[2]} = {bg=gray!25}}%
\toprule
        Benchmark & Used by & Source & Agentic  \\ 
\midrule
        SimpleQA & OpenAI & Introducing GPT-4.5 \cite{gpt-4.5} & \xmark  \\ 
        SWE-Bench Verified & OpenAI & Introducing GPT-4.5 \cite{gpt-4.5} & \cmark  \\ 
        GPQA & OpenAI & Introducing GPT-4.5 \cite{gpt-4.5} & \xmark  \\ 
        AIME ‘24 & OpenAI & Introducing GPT-4.5 \cite{gpt-4.5} & \xmark  \\ 
        MMMLU & OpenAI & Introducing GPT-4.5 \cite{gpt-4.5} & \xmark  \\ 
        MMMU & OpenAI & Introducing GPT-4.5 \cite{gpt-4.5} & \xmark  \\ 
        SWE-Lancer Diamond & OpenAI & Introducing GPT-4.5 \cite{gpt-4.5} & \cmark  \\ 
        GAIA & OpenAI & Introducing deep research \cite{deep-research} & \cmark  \\ 
        FrontierMath & OpenAI & OpenAI o3-mini \cite{o3-mini} & \cmark  \\ 
        Codeforces & OpenAI & OpenAI o3-mini \cite{o3-mini} & \cmark  \\ 
        LiveBench Coding & OpenAI & OpenAI o3-mini \cite{o3-mini} & \cmark  \\ 
        MMLU & OpenAI & OpenAI o3-mini \cite{o3-mini} & \xmark  \\ 
        Math & OpenAI & OpenAI o3-mini \cite{o3-mini} & \xmark  \\ 
        MGSM & OpenAI & OpenAI o3-mini \cite{o3-mini} & \xmark  \\ 
        OSWorld & OpenAI & Computer-Using Agent \cite{computer-user} & \cmark  \\ 
        WebArena & OpenAI & Computer-Using Agent \cite{computer-user} & \cmark  \\ 
        WebVoyager & OpenAI & Computer-Using Agent \cite{computer-user} & \cmark  \\ 
        HumanEval & OpenAI & OpenAI o1-mini \cite{o1-mini} & \xmark  \\ 
        MATH-500 & OpenAI & OpenAI o1-mini \cite{o1-mini} & \cmark  \\ 
        DROP & OpenAI & GPT-4o mini: advancing cost-efficient intelligence \cite{4o-mini} & \xmark  \\ 
        MathVista & OpenAI & GPT-4o mini: advancing cost-efficient intelligence \cite{4o-mini} & \cmark  \\ 
        RE-Bench & OpenAI & GPT-4o System Card \cite{4o} & \cmark  \\ 
        MedQA & OpenAI & GPT-4o System Card \cite{4o} & \xmark  \\ 
        MedMCQA & OpenAI & GPT-4o System Card \cite{4o} & \xmark  \\ 
        ProtocolQA & OpenAI & OpenAI o1 System Card \cite{o1} & \xmark  \\ 
        BioLP-Bench & OpenAI & OpenAI o1 System Card \cite{o1} & \xmark  \\ 
        MLE-bench & OpenAI & OpenAI o1 System Card \cite{o1} & \cmark  \\ 
        Tau-bench & Anthropic & Claude 3.7 Sonnet and Claude Code \cite{claude-3.7} & \cmark  \\ 
        BIG-Bench-Hard & Anthropic & Claude 3.5 Sonnet \cite{claude-3.5} & \xmark  \\ 
        IF-Eval & Deepseek & Introducing DeepSeek-V3 \cite{deepseek-V3} & \xmark  \\ 
        FRAMES & Deepseek & Introducing DeepSeek-V3 \cite{deepseek-V3} & \xmark  \\ 
        LongBench v2 & Deepseek & Introducing DeepSeek-V3 \cite{deepseek-V3} & \xmark  \\ 
        Aider-Edit & Deepseek & Introducing DeepSeek-V3 \cite{deepseek-V3} & \cmark  \\ 
        Aider-Polyglot & Deepseek & Introducing DeepSeek-V3 \cite{deepseek-V3} & \cmark  \\ 
        CNMO 2024 & Deepseek & Introducing DeepSeek-V3 \cite{deepseek-V3} & \cmark  \\ 
        CLUEWSC & Deepseek & Introducing DeepSeek-V3 \cite{deepseek-V3} & \xmark  \\ 
        C-Eval & Deepseek & Introducing DeepSeek-V3 \cite{deepseek-V3} & \xmark  \\ 
        C-SimpleQA & Deepseek & Introducing DeepSeek-V3 \cite{deepseek-V3} & \xmark  \\ 
        LOFT (128k) & xAI & Grok 3 Beta --- The Age of Reasoning Agents \cite{grok-3} & \xmark  \\ 
        EgoSchema & xAI & Grok 3 Beta --- The Age of Reasoning Agents \cite{grok-3} & \xmark  \\ 
        DocVQA & xAI & Grok-2 Beta Release \cite{grok-2} & \xmark  \\ 
        ChartQA & Meta & Llama 3.2: Revolutionizing edge AI and vision with open, customizable models \cite{llama-3.2} & \xmark  \\ 
        AI2 Diagram & Meta & Llama 3.2: Revolutionizing edge AI and vision with open, customizable models \cite{llama-3.2} & \xmark  \\ 
        VQAv2 & Meta & Llama 3.2: Revolutionizing edge AI and vision with open, customizable models \cite{llama-3.2} & \xmark  \\ 
        Open-rewrite eval & Meta & Llama 3.2: Revolutionizing edge AI and vision with open, customizable models \cite{llama-3.2} & \xmark  \\ 
        TLDR9+ & Meta & Llama 3.2: Revolutionizing edge AI and vision with open, customizable models \cite{llama-3.2} & \xmark  \\ 
        BFCL V2 & Meta & Llama 3.2: Revolutionizing edge AI and vision with open, customizable models \cite{llama-3.2} & \xmark  \\ 
        Nexus & Meta & Llama 3.2: Revolutionizing edge AI and vision with open, customizable models \cite{llama-3.2} & \xmark  \\ 
        ARC Challenge & Meta & Llama 3.2: Revolutionizing edge AI and vision with open, customizable models \cite{llama-3.2} & \xmark  \\ 
        Hellaswag & Meta & Llama 3.2: Revolutionizing edge AI and vision with open, customizable models \cite{llama-3.2} & \xmark  \\ 
        InfiniteBench & Meta & Llama 3.2: Revolutionizing edge AI and vision with open, customizable models \cite{llama-3.2} & \xmark  \\ 
        NIH/Multi-needle & Meta & Llama 3.2: Revolutionizing edge AI and vision with open, customizable models \cite{llama-3.2} & \xmark  \\ 
        ZeroScrolls & Meta & Introducing Llama 3.1: Our most capable models to date \cite{llama-3.1} & \xmark  \\ 
        Bird-Bench & Google & Gemini 2.0 is now available to everyone \cite{gemini-2.0} & \cmark  \\ 
        FACTS Grounding & Google & Gemini 2.0 is now available to everyone \cite{gemini-2.0} & \xmark  \\ 
        HiddenMath & Google & Gemini 2.0 is now available to everyone \cite{gemini-2.0} & \cmark  \\ 
        MRCR & Google & Gemini 2.0 is now available to everyone \cite{gemini-2.0} & \xmark  \\ 
        CoVoST2 & Google & Gemini 2.0 is now available to everyone \cite{gemini-2.0} & \xmark  \\ 
        MBPP & Mistral & Mistral Large 2 \cite{mixtral-large} & \xmark  \\ 
        MT-Bench & Mistral & Mistral Large 2 \cite{mixtral-large} & \xmark  \\ 
        Wild Bench & Mistral & Mistral Large 2 \cite{mixtral-large} & \xmark  \\ 
        Arena Hard & Mistral & Mistral Large 2 \cite{mixtral-large} & \xmark  \\ 
        BBH & Amazon & The Amazon Nova family of models: Technical report and model card \cite{amazon-nova} & \xmark  \\ 
        ARC-C & Amazon & The Amazon Nova family of models: Technical report and model card \cite{amazon-nova} & \xmark  \\ 
        ChartQA & Amazon & The Amazon Nova family of models: Technical report and model card \cite{amazon-nova} & \xmark  \\ 
        Doc VQA & Amazon & The Amazon Nova family of models: Technical report and model card \cite{amazon-nova} & \xmark  \\ 
        VATEX & Amazon & The Amazon Nova family of models: Technical report and model card \cite{amazon-nova} & \xmark  \\ 
        Text VQA & Amazon & The Amazon Nova family of models: Technical report and model card \cite{amazon-nova} & \xmark  \\ 
        Ego Schema & Amazon & The Amazon Nova family of models: Technical report and model card \cite{amazon-nova} & \xmark  \\ 
        VisualWebBench & Amazon & The Amazon Nova family of models: Technical report and model card \cite{amazon-nova} & \xmark  \\ 
        NN-Mind2Web & Amazon & The Amazon Nova family of models: Technical report and model card \cite{amazon-nova} & \xmark  \\ 
        GroundUI-1K & Amazon & The Amazon Nova family of models: Technical report and model card \cite{amazon-nova} & \xmark  \\ 
        SQuALITY & Amazon & The Amazon Nova family of models: Technical report and model card \cite{amazon-nova} & \xmark  \\ 
        LVBench & Amazon & The Amazon Nova family of models: Technical report and model card \cite{amazon-nova} & \xmark  \\ 
        FinQA & Amazon & The Amazon Nova family of models: Technical report and model card \cite{amazon-nova} & \xmark  \\ 
        CRAG & Amazon & The Amazon Nova family of models: Technical report and model card \cite{amazon-nova} & \xmark  \\ 
        Kernel-Bench & DL4C & KernelBench: Can LLMs Write Efficient GPU Kernels? \cite{ouyang2025kernelbench} & \cmark \\
        \bottomrule
    \end{longtblr}
    }

%% file: tex/app-sources.tex
\section{Sources of the Checks in \name} \label{sec:app-source}
In Table \ref{tab:sources}, we show the detail construction process of \name by 
listing the sources of each check proposed in \name. We synthesized the insights
from the following aspects
\begin{enumerate}[leftmargin=*]
    \item Our experience of developing agentic benchmarks.
    \item Best practices in existing agentic benchmarks (Table \ref{tab:all-benchmarks}).
    \item Lessons learned from issues of existing agentic benchmarks.
    \item Domain-specific suggestions when we apply well-established 
    techniques as evaluation methods.
\end{enumerate}

\input{figures/tab_checklist_sources.tex}

%% file: figures/tab_checklist_sources.tex
{\scriptsize
\begin{longtblr}[
    caption = {Sources of items in \name},
    label = {tab:sources} ]{colspec={Q[c]Q[l]Q[l]Q[l]},row{1} = {font=\bfseries}, row{odd[2]} = {bg=gray!25}}%
\toprule
Question & Existing Best Practice & Lessons Learned & Domain-Specific Suggestions \\
\midrule
O.a.1   & {\citet{mialon2023gaia},\\\citet{zhouwebarena}}    &          & \\
O.a.2   & {\citet{mialon2023gaia},\\\citet{zhouwebarena}}    & \citet{zhouwebarena}          & \\
O.b.1   & \citet{mialon2023gaia}    & \citet{zhouwebarena}          & \\
O.b.2   &                           & \citet{zhouwebarena,yaotau}    & \\
O.b.3   & \citet{zhouwebarena}      & \citet{yaotau}                & \\
O.c.1   & \citet{he2024webvoyager}  &                               & \citet{ziems2024can} \\
O.d.1   & \citet{swebench-verified} & \citet{utboost,jimenez2024swe} & \\
O.d.2   & & & \citet{zhu1997software} \\
O.e.1   &                           & \citet{ouyang2025kernelbench} & \citet{zhu2022fuzzing} \\
O.e.2   &                           & \citet{ouyang2025kernelbench} & \citet{zhu2022fuzzing}\\
O.e.3   & \citet{metr-kernel}       & & \\
O.f.1   &                           &                               & \citet{ricca2001analysis} \\
O.f.2   &                           &                               & \citet{parry2021survey} \\
O.g.1   & {\citet{yaotau},\\\citet{zhouwebarena},\\\citet{xie2024osworld}} & & \\
O.g.2   & \citet{yaotau}             & \citet{xie2024osworld} & \\
O.g.3   &                           & \citet{yaotau} & \\
O.h.1   & \citet{mialon2023gaia}     & {\citet{key},\\\citet{lightman2023let}}    & \\
O.h.2   & \citet{glazer2024frontiermath} & & \\
O.i.1   & \citet{chan2024mle}        & & \\
T.1    & {\citet{miserendino2025swe},\\\citet{li2023can}}          & & \\
T.2    & \citet{kapoor2024ai}       & \citet{zhouwebarena} & \\
T.3    & \citet{zhouwebarena}       & \citet{zhu2025cve} & \\
T.4    & {\citet{miserendino2025swe},\\\citet{yaotau},\\\citet{jimenez2024swe}} & \citet{lange2025ai} & \\
T.5    & \citet{zhang2024cybench}  & \citet{miserendino2025swe} & \\
T.6    & & {\citet{wretblad2024understanding},\\\citet{pourreza2023evaluating},\\\citet{li2023can}} & \\
T.7    & {\citet{zhang2024cybench},\\\citet{zhu2025cve},\\\cite{xie2024osworld}} & & \\
T.8    & {\citet{zhang2024cybench},\\\citet{zhu2025cve}} & \citet{li2023can} & \\
T.9    & & {\citet{lange2025ai},\\\citet{miserendino2025swe}} & \\
R.1   & All benchmarks in Table \ref{tab:benchmarks}. & & \\
R.2   & {All benchmarks in Table \ref{tab:benchmarks}\\ except \gaia.} & & \\
R.3   & {\citet{chan2024mle},\\\citet{miserendino2025swe},\\\cite{li2023can}} & & \citet{zhou2023don} \\
R.4   & \citet{white2024livebench} & & \\
R.5   & \citet{kapoor2024ai} & & \\
R.6   & All benchmarks in Table \ref{tab:benchmarks}. & & \\
R.7   & \citet{chan2024mle,yaotau} & & \\
R.8   & {\citet{miserendino2025swe},\\\citet{chan2024mle}} & & \\
R.9   & \citet{yaotau} & & \\
R.10   & & & {\citet{dorner2024don},\\\citet{reuelbetterbench}} \\
R.11   & & & {\citet{hothorn2005design},\\\citet{dorner2024don}} \\
R.12  & {\citet{cao2025should,xie2024osworld},\\\citet{zhang2024cybench}} & & \\
R.13  & & \citet{yaotau} & \\
\bottomrule
\end{longtblr}
}

%% file: tex/app-assessment.tex
\section{Assessment Reports} \label{sec:app-reports}

In this section we provide detailed assessment reports for all ten benchmarks. 
Each report’s caption specifies the corresponding paper and codebase evaluated.

\input{figures/tab_full_assessment.tex}

%% file: figures/tab_full_assessment.tex
{\scriptsize
\begin{longtblr}[
    caption = {Assessment Report of SWE-Bench-Lancer (\href{https://arxiv.org/pdf/2502.12115}{paper}, \href{https://github.com/openai/SWELancer-Benchmark?tab=readme-ov-file}{code})},
    label = {tab:sources} ]{colspec={Q[c]Q[l]Q[l]Q[l]},row{1} = {font=\bfseries}, row{odd[2]} = {bg=gray!25}}%
\toprule
Check & Score & Reason \\
\midrule
\textbf{O.d.1} & 1 & {As discussed in Section 1 of the paper, the benchmark uses a set of test cases that are verified for correctness\\ and quality by human experts.} \\
\textbf{O.d.2} & 0 & The benchmark does not use objective metrics to measure the quality of test cases. \\
\textbf{O.f.2} & 1 & As discussed in Section 1, the end-to-end testing is designed to simulate the entire user workflow. \\
\textbf{O.f.3} & 0 & {The test cases use hard-coded timeouts, which may lead to non-deterministic results if the system is slow or\\ unresponsive.} \\
\textbf{T.1} & 1 & The package dependencies are specified in the repository of each task. \\
\textbf{T.2} & 1 & The benchmark does not require any external APIs. \\
\textbf{T.3} & 1 & The benchmark does not require any external APIs. \\
\textbf{T.4} & 1 & The benchmark uses docker containers to isolate the environment, and the state is cleared between runs. \\
\textbf{T.5} & 0 & {The agent can access the file system where the test cases are stored, which may lead to the agent accessing the\\ ground truth information.} \\
\textbf{T.6} & 1 & The environment setup is static and does not change over time. \\
\textbf{T.7} & 1 & The ground-truth test cases are taken from GitHub repositories, which are verified by expert developers. \\
\textbf{T.8} & 1 & Each task represents a real-world software issue with a corresponding patch, which are solvable by the agent. \\
\textbf{T.9} & 1 & The benchmark uses existing patches as ground truth, which can be considered as an Oracle solver. \\
\textbf{T.10} & 0 & {The benchmark does not handle the isolation between the agent and test cases properly. The test cases are stored\\ not only in a file system that the agent can access, but also in a ZIP file that agent can read the directory structure and update files.} \\
\textbf{R.1} & 1 & The benchmark is open-sourced and available on GitHub. \\
\textbf{R.2} & 1 & The benchmark provides an open-source evaluation harness for users. \\
\textbf{R.3} & 1 & The benchmark maintains a private test set. \\
\textbf{R.4} & 0 & The report does not discuss any measures or plans for consistent update. \\
\textbf{R.5} & 1 & Such a relationship is clearly stated in Section 2 of the paper. \\
\textbf{R.6} & 1 & As shown in Section 3, the benchmark is designed to evaluate the LLM model. \\
\textbf{R.7} & 1 & The benchmark uses end-to-end testing to mitigate grader hacking. \\
\textbf{R.8} & 1 & The benchmark discusses the potential impact of grader hacking in Section 1 and Appendix A.7. \\
\textbf{R.9} & 0 & The benchmark does not include any quantitative analysis to assess the impact of grader hacking. \\
\textbf{R.10} & 0 & The benchmark does not report any metrics about statistical significance. \\
\textbf{R.11} & 0 & The benchmark does not provide any guidance on interpreting results with eval flaws. \\
\textbf{R.12} & 0 & The benchmark does not report results of non-AI baselines. \\
\textbf{R.13} & 0 & The benchmark does not report results of trivial agents. \\
\bottomrule
\end{longtblr}
}
{\scriptsize
\begin{longtblr}[
    caption = {Assessment Report of Bird-Bench (\href{https://arxiv.org/pdf/2305.03111}{paper}, \href{https://github.com/AlibabaResearch/DAMO-ConvAI/tree/main/bird}{code})},
    label = {tab:sources} ]{colspec={Q[c]Q[l]Q[l]Q[l]},row{1} = {font=\bfseries}, row{odd[2]} = {bg=gray!25}}%
\toprule
Check & Score & Reason \\
\midrule
\textbf{O.d.1} & 1 & {As discussed in Section 3.4 of the paper, the validity of the database is verified by executing the ground-truth\\ query.} \\
\textbf{O.d.2} & 0 & {The paper does not use objective metrics to measure the usefulness and completeness of the database or\\ ground-truth queries.} \\
\textbf{O.f.2} & 0 & The paper does not provide any information about the coverage of the database or ground-truth queries. \\
\textbf{O.f.3} & 1 & {Executing SQL queries on a database is deterministic, and the paper does not mention any non-deterministic\\ behavior.} \\
\textbf{T.1} & 1 & The task instruction in Figure 9 specifies the SQL language is SQLite. \\
\textbf{T.2} & 1 & No external API is required for the evaluation of the benchmark. \\
\textbf{T.3} & 1 & No external API is required for the evaluation of the benchmark. \\
\textbf{T.4} & 0 & {Database file is neither opened in a read-only mode nor re-initialized between runs. This may lead to unexpected\\ data manipulation by the agent.} \\
\textbf{T.5} & 1 & Agent cannot access the host file system. \\
\textbf{T.6} & 1 & The environment setup is static and does not change over time. \\
\textbf{T.7} & 0 & {As discussed in Section 3.4 of the paper, the correctness of the query is not fully verified, especially for the\\ SQL queries that two annotators reach a consensus on.} \\
\textbf{T.8} & 0 & The ambiguity of the SQL queries is not fully verified. \\
\textbf{T.9} & 0 & The Benchmark does not include an Oracle solver that can automatically solve all text-to-SQL tasks. \\
\textbf{T.10} & 1 & No vulnerabilities are found in the implementation of the benchmark. \\
\textbf{R.1} & 1 & The benchmark is open-sourced and available on GitHub. \\
\textbf{R.2} & 1 & The benchmark provides an open-source evaluation harness for users. \\
\textbf{R.3} & 1 & The benchmark has a private test set. \\
\textbf{R.4} & 0 & The benchmark does not discuss any measures or plans for consistent update. \\
\textbf{R.5} & 1 & It is clearly discussed in Section 2 of the paper. \\
\textbf{R.6} & 1 & It is clearly discussed in Section 2 of the paper. \\
\textbf{R.7} & 0 & No efforts are made to mitigate errors when both annotators make the same mistake. \\
\textbf{R.8} & 0 & The paper does not discuss the potential impact of annotation errors. \\
\textbf{R.9} & 0 & The paper does not analyze the quantitative impact of annotation errors. \\
\textbf{R.10} & 0 & The paper does not report any metrics about statistical significance. \\
\textbf{R.11} & 0 & The paper does not provide any guidance on interpreting results with eval flaws. \\
\textbf{R.12} & 1 & The paper reports the results of human experts. \\
\textbf{R.13} & 0 & The paper does not report the results of any trivial agents. \\
\bottomrule
\end{longtblr}
}
{\scriptsize
\begin{longtblr}[
    caption = {Assessment Report of CyBench (\href{https://arxiv.org/pdf/2408.08926}{paper}, \href{https://github.com/andyzorigin/cybench}{code})},
    label = {tab:sources} ]{colspec={Q[c]Q[l]Q[l]Q[l]},row{1} = {font=\bfseries}, row{odd[2]} = {bg=gray!25}}%
\toprule
Check & Score & Reason \\
\midrule
\textbf{O.h.1} & 1 & The specific format required for the answer is provided in the task description. \\
\textbf{O.h.2} & 1 & The ground truth is complex enough to prevent trivial guessing. \\
\textbf{T.1} & 1 & Agents are granted access to using all tools. The versions of tools can be checked by the agent. \\
\textbf{T.2} & 1 & The benchmark does not require any external APIs. \\
\textbf{T.3} & 1 & The benchmark does not require any external APIs. \\
\textbf{T.4} & 1 & The benchmark uses docker containers to isolate the environment, and the state is cleared between runs. \\
\textbf{T.5} & 1 & The agent cannot directly access the container which contains the ground truth. \\
\textbf{T.6} & 1 & The environment setup is static and does not change over time. \\
\textbf{T.7} & 1 & As shown in Section 3.3 of the paper, the ground truth is verified by human. \\
\textbf{T.8} & 1 & As shown in Section 3.3 of the paper, each task is verified to be solvable. \\
\textbf{T.9} & 1 & {As shown in Section 3.3 of the paper, the benchmark includes an Oracle solver that can automatically solve all\\ tasks.} \\
\textbf{T.10} & 1 & No vulnerabilities are found in the implementation of the benchmark. \\
\textbf{R.1} & 1 & The benchmark is open-sourced and available on GitHub. \\
\textbf{R.2} & 1 & The benchmark provides an open-source evaluation harness for users. \\
\textbf{R.3} & 0 & The benchmark does not contain measures to prevent data contamination. \\
\textbf{R.4} & 0 & The report does not discuss plans to consistently update tasks over time. \\
\textbf{R.5} & 1 & Such a relationship is clearly stated in Section 1 of the paper. \\
\textbf{R.6} & 1 & As shown in Section 1, the benchmark is designed to evaluate both agent frameworks and LLM models. \\
\textbf{R.7} & 1 & Annotation flaws are mitigated by developing verifiable tasks. \\
\textbf{R.8} & 1 & No unavoidable flaws are identified in the benchmark. \\
\textbf{R.9} & 1 & No unavoidable flaws are identified in the benchmark. \\
\textbf{R.10} & 0 & The report does not include any metrics about statistical significance. \\
\textbf{R.11} & 1 & No evaluation flaws are identified in the benchmark. \\
\textbf{R.12} & 1 & Human performance is reported in Section 5 of the paper. \\
\textbf{R.13} & 0 & The report does not report results of trivial agents. \\
\bottomrule
\end{longtblr}
}
{\scriptsize
\begin{longtblr}[
    caption = {Assessment Report of SWE-Bench-Verified (\href{https://arxiv.org/pdf/2310.06770}{paper}, \href{https://github.com/SWE-bench/SWE-bench}{code})},
    label = {tab:sources} ]{colspec={Q[c]Q[l]Q[l]Q[l]},row{1} = {font=\bfseries}, row{odd[2]} = {bg=gray!25}}%
\toprule
Check & Score & Reason \\
\midrule
\textbf{O.d.1} & 1 & Test cases are directly taken from GitHub repositories, and the paper does not mention any verification process. \\
\textbf{O.d.2} & 0 & The paper does not use objective metrics to measure quality of test cases. \\
\textbf{T.1} & 1 & The versions of package dependencies are specified in the repository. \\
\textbf{T.2} & 1 & The benchmark does not require any external APIs. \\
\textbf{T.3} & 1 & The benchmark does not require any external APIs. \\
\textbf{T.4} & 1 & The benchmark uses docker containers to isolate the environment, and the state is cleared between runs. \\
\textbf{T.5} & 1 & The agent cannot access the host file system, and the ground truth is not accessible to the agent. \\
\textbf{T.6} & 1 & The environment setup is static and does not change over time. \\
\textbf{T.7} & 1 & The ground-truth patches are taken from GitHub repositories, which is verified by expert developers. \\
\textbf{T.8} & 1 & Each task represents a real-world GitHub issue and a corresponding pull request, which are solvable by the agent. \\
\textbf{T.9} & 1 & Pull requests from GitHub are used as ground truth, which can be considered as an Oracle solver. \\
\textbf{T.10} & 1 & No vulnerabilities are found in the implementation of the benchmark, and the evaluation process is secure. \\
\textbf{R.1} & 1 & The benchmark is open-sourced and available on GitHub. \\
\textbf{R.2} & 1 & The benchmark provides an open-source evaluation harness for users. \\
\textbf{R.3} & 0 & The benchmark does not discuss measures to prevent data contamination. \\
\textbf{R.4} & 0 & The benchmark does not discuss plans to consistently update tasks over time. \\
\textbf{R.5} & 1 & Such a relationship is clearly stated in Section 2 of the paper. \\
\textbf{R.6} & 1 & {The benchmark is designed to evaluate both the model and the agent framework, as discussed in Section 5 of the\\ paper.} \\
\textbf{R.7} & 0 & The benchmark does not discuss any efforts to prevent, identify, and correct flaws. \\
\textbf{R.8} & 0 & The benchmark does not discuss the potential impact of unavoidable flaws. \\
\textbf{R.9} & 0 & The benchmark does not include quantitative analysis to assess the impact of unavoidable flaws. \\
\textbf{R.10} & 0 & The report does not include any metrics about statistical significance. \\
\textbf{R.11} & 0 & The benchmark does not provide any guidance on interpreting results with eval flaws. \\
\textbf{R.12} & 0 & The benchmark does not report results of non-AI baselines. \\
\textbf{R.13} & 0 & The benchmark does not report results of trivial agents. \\
\bottomrule
\end{longtblr}
}
{\scriptsize
\begin{longtblr}[
    caption = {Assessment Report of $\\tau$-Bench (\href{https://arxiv.org/pdf/2406.12045}{paper}, \href{https://github.com/sierra-research/tau-bench}{code})},
    label = {tab:sources} ]{colspec={Q[c]Q[l]Q[l]Q[l]},row{1} = {font=\bfseries}, row{odd[2]} = {bg=gray!25}}%
\toprule
Check & Score & Reason \\
\midrule
\textbf{O.a.1} & 1 & The benchmark uses minimal expressions for substring matching, which is robust to variations in the input. \\
\textbf{O.a.2} & 1 & The benchmark uses minimal expressions for substring matching, which is robust to redundant words in the input. \\
\textbf{O.b.1} & 0 & The benchmark does not specify how negation modifiers are handled, which may lead to incorrect evaluations. \\
\textbf{O.b.2} & 0 & {The benchmark does not specify how it handles systematic listing of all possible answers, which may lead to\\ incorrect evaluations.} \\
\textbf{O.b.3} & 0 & A part of tasks has empty ground truth, which may lead to guessing. \\
\textbf{O.g.1} & 1 & The database after successful completion of a task is unique and includes all states. \\
\textbf{O.g.2} & 1 & The state of the database is the only environment state, and it is checked for both relevant and irrelevant parts. \\
\textbf{O.g.3} & 0 & A part of tasks has empty ground truth, which may lead to trivial state modifications. \\
\textbf{T.1} & 1 & The benchmark does not use external tools. \\
\textbf{T.2} & 1 & The benchmark does not use external APIs. \\
\textbf{T.3} & 1 & The benchmark does not use external APIs. \\
\textbf{T.4} & 1 & Residual data or state are fully cleared between runs by re-initializing the database. \\
\textbf{T.5} & 1 & Agents has no access to the file system. \\
\textbf{T.6} & 1 & The environment setup is static and does not change over time. \\
\textbf{T.7} & 1 & As shown in Section 4 of the paper, the ground truth is manually verified. \\
\textbf{T.8} & 1 & As shown in Section 4 of the paper, each task is verified to be solvable by the agent. \\
\textbf{T.9} & 1 & The benchmark provides a reference task solution that can be used as an Oracle solver. \\
\textbf{T.10} & 1 & No vulnerabilities are found in the implementation of the benchmark, and the evaluation process is secure. \\
\textbf{R.1} & 1 & The benchmark is open-sourced and available on GitHub. \\
\textbf{R.2} & 1 & The benchmark provides an open-source evaluation harness for users. \\
\textbf{R.3} & 0 & The benchmark does not discuss measures to prevent data contamination. \\
\textbf{R.4} & 0 & The report does not discuss plans to consistently update tasks over time. \\
\textbf{R.5} & 1 & Such a relationship is clearly stated in Section 3 of the paper. \\
\textbf{R.6} & 1 & {As discussed in Section 5 of the paper, the benchmark is designed to evaluate both the model and the agent\\ framework.} \\
\textbf{R.7} & 1 & Appendix A of the paper shows the efforts taken to detect annotation errors. \\
\textbf{R.8} & 1 & Section 6 discusses the potential impact of unavoidable flaws, although these discussions are not sufficient. \\
\textbf{R.9} & 0 & The report does not include quantitative analysis to assess the impact of unavoidable flaws. \\
\textbf{R.10} & 0 & The report does not include any metrics about statistical significance. \\
\textbf{R.11} & 0 & The report does not provide any guidance on interpreting results with eval flaws. \\
\textbf{R.12} & 0 & The report does not report results of non-AI baselines. \\
\textbf{R.13} & 0 & The report does not report results of trivial agents. \\
\bottomrule
\end{longtblr}
}
{\scriptsize
\begin{longtblr}[
    caption = {Assessment Report of MLE-Bench (\href{https://arxiv.org/pdf/2410.07095}{paper}, \href{https://github.com/openai/mle-bench/tree/main}{code})},
    label = {tab:sources} ]{colspec={Q[c]Q[l]Q[l]Q[l]},row{1} = {font=\bfseries}, row{odd[2]} = {bg=gray!25}}%
\toprule
Check & Score & Reason \\
\midrule
\textbf{O.I.1} & 1 & As described in Section 2.2, the benchmark uses leaderboard positions as a metric, which is not easily exploitable. \\
\textbf{T.1} & 0 & The prompt does not specify the versions of important tools, such as Python and Pytorch. \\
\textbf{T.2} & 1 & The benchmark does not require any external APIs, and all required tools are accessible to the agent. \\
\textbf{T.3} & 1 & {The benchmark does not require any external APIs, and the evaluation process does not depend on any external\\ resources.} \\
\textbf{T.4} & 1 & There are no residual data or state between runs, as the evaluation is performed in a clean environment. \\
\textbf{T.5} & 1 & {The submission process is isolated from the agent's environment, and the agent cannot access any ground truth\\ information.} \\
\textbf{T.6} & 1 & The environment setup is static and does not change over time. \\
\textbf{T.7} & 1 & {The benchmark uses ground truth data from Kaggle, which is a widely used and reliable source for benchmark\\ datasets.} \\
\textbf{T.8} & 1 & The benchmark uses previous challenges from Kaggle, which are proven to be solvable with ML algorithms. \\
\textbf{T.9} & 1 & Any solution on Kaggle can be considered an Oracle solver. \\
\textbf{T.10} & 1 & No vulnerabilities are found in the implementation of the benchmark, and the evaluation process is secure. \\
\textbf{R.1} & 1 & The benchmark is open-sourced and available on GitHub. \\
\textbf{R.2} & 1 & The benchmark provides an open-source evaluation harness for users. \\
\textbf{R.3} & 1 & The benchmark design experiments to measure data contamination and agent plagiarism. \\
\textbf{R.4} & 1 & Future plan on regularly update the benchmark with new Kaggle challenges is discussed in Section 6 \\
\textbf{R.5} & 1 & Such a relationship is clearly stated in Section 2. \\
\textbf{R.6} & 1 & As shown in Section 3, the benchmark is designed to evaluate both the model and the agent framework. \\
\textbf{R.7} & 1 & The paper discusses the efforts taken to detect cheating in Appendix A.5. \\
\textbf{R.8} & 1 & The paper discusses the potential impact of unavoidable flaws in Section 4. \\
\textbf{R.9} & 1 & The paper includes quantitative analysis to assess the impact of unavoidable flaws in Appendix A.5. \\
\textbf{R.10} & 1 & The paper reports metrics about statistical significance in Section 3.3. \\
\textbf{R.11} & 1 & No significant flaws are found in the evaluation process. \\
\textbf{R.12} & 1 & The benchmark directly compares the performance of agents with human experts in the Kaggle challenge submissions. \\
\textbf{R.13} & 0 & The benchmark does not report results of trivial agents. \\
\bottomrule
\end{longtblr}
}
{\scriptsize
\begin{longtblr}[
    caption = {Assessment Report of WebArena (\href{https://arxiv.org/pdf/2307.13854}{paper}, \href{https://github.com/web-arena-x/webarena}{code})},
    label = {tab:sources} ]{colspec={Q[c]Q[l]Q[l]Q[l]},row{1} = {font=\bfseries}, row{odd[2]} = {bg=gray!25}}%
\toprule
Check & Score & Reason \\
\midrule
\textbf{O.a.1} & 1 & {As discussed in Section 3.2 of the paper, the benchmark expects the response to follow a standardized format,\\ which is robust to variations in the input.} \\
\textbf{O.a.2} & 1 & {As discussed in Section 3.2 of the paper, the benchmark expects the response to follow a standardized format,\\ which is robust to redundant words in the input.} \\
\textbf{O.b.1} & 0 & The benchmark does not handle negation modifiers, which may lead to incorrect evaluations. \\
\textbf{O.b.2} & 0 & {The benchmark does not specify how it handles systematic listing of all possible answers, which may lead to\\ incorrect evaluations.} \\
\textbf{O.b.3} & 0 & The ground truth is NULL for a part of tasks, which may lead to guessing. \\
\textbf{O.c.1} & 1 & The accuracy of the judge is quantitatively evaluated in Appendix A.8 of the paper. \\
\textbf{O.c.2} & 0 & {The benchmark does not handle adversarial inputs and reward hacking in LLM-as-a-Judge, which may lead to incorrect\\ evaluations.} \\
\textbf{O.g.1} & 1 & The ground truth includes all states achievable after success, as discussed in Section 3.2 of the paper. \\
\textbf{O.g.2} & 0 & {The state check only considers relevant states (e.g., achieved by using a locator as discussed in Section 3.2),\\ which may lead to incorrect evaluations.} \\
\textbf{O.g.3} & 1 & {As demonstrated in Section 3.2 of the paper, the ground truth is a modification of the underlying database, which\\ is complex enough to prevent trivial state modifications.} \\
\textbf{T.1} & 1 & The benchmark does not use tools that require version specification. \\
\textbf{T.2} & 0 & {The benchmark requires an external API (e.g., a clone of Reddit website) that is not can be inaccessible to agents\\ during evaluation due to rate limit.} \\
\textbf{T.3} & 0 & {The evaluation process does not handle errors appropriately if the API becomes inaccessible, which may lead to\\ incorrect evaluations.} \\
\textbf{T.4} & 1 & The benchmark uses docker containers to isolate the environment, and the state is cleared between runs. \\
\textbf{T.5} & 1 & The agent has no access to the file system where the ground truth is stored. \\
\textbf{T.6} & 1 & The environment setup is static and does not change over time. \\
\textbf{T.7} & 0 & {As mentioned in Section 3.2, the ground truth is annotated by two human annotators. However, there isn't a\\ mechanism to verify or guarantee the correctness of the annotations.} \\
\textbf{T.8} & 0 & The ambiguity of the tasks is not fully verified or tested, which may lead to incorrect evaluations. \\
\textbf{T.9} & 0 & The benchmark does not include an Oracle solver that can automatically solve all tasks. \\
\textbf{T.10} & 0 & A do-nothing agent can pass 4.4
\textbf{R.1} & 1 & The benchmark is open-sourced and available on GitHub. \\
\textbf{R.2} & 1 & The benchmark provides an open-source evaluation harness for users. \\
\textbf{R.3} & 0 & The benchmark does not discuss measures to prevent data contamination. \\
\textbf{R.4} & 0 & The benchmark does not discuss plans to consistently update tasks over time. \\
\textbf{R.5} & 1 & Such a relationship is clearly stated in Section 2.1 of the paper. \\
\textbf{R.6} & 1 & As shown in Section 5, the benchmark is designed to evaluate LLM models. \\
\textbf{R.7} & 1 & Efforts to evaluate LLM-as-a-Judge are discussed in Appendix A.8 of the paper. \\
\textbf{R.8} & 0 & The report does not discuss the potential impact of unavoidable flaws. \\
\textbf{R.9} & 0 & The report does not include quantitative analysis to assess the impact of unavoidable flaws. \\
\textbf{R.10} & 0 & The report does not include any metrics about statistical significance. \\
\textbf{R.11} & 0 & The report does not provide any guidance on interpreting results with eval flaws. \\
\textbf{R.12} & 1 & The human performance is reported in appendix A.5. \\
\textbf{R.13} & 0 & The report does not report results of trivial agents. \\
\bottomrule
\end{longtblr}
}
{\scriptsize
\begin{longtblr}[
    caption = {Assessment Report of GAIA (\href{https://arxiv.org/pdf/2311.12983}{paper}, \href{https://huggingface.co/gaia-benchmark}{code})},
    label = {tab:sources} ]{colspec={Q[c]Q[l]Q[l]Q[l]},row{1} = {font=\bfseries}, row{odd[2]} = {bg=gray!25}}%
\toprule
Check & Score & Reason \\
\midrule
\textbf{O.h.1} & 1 & {As discussed in Section 3.2 of the paper, the specific format required for the answer is provided in the task\\ description.} \\
\textbf{O.h.2} & 1 & The ground truth is complex enough to prevent trivial guessing. \\
\textbf{T.1} & 0 & The version of tools (e.g., Python and website) is not specified in the paper. \\
\textbf{T.2} & 0 & The rate limit of the API is not specified in the paper, which may lead to incorrect evaluations. \\
\textbf{T.3} & 0 & {The benchmark does not provide a reference harness for handling errors, which may lead to inconsistent evaluations\\ across different users.} \\
\textbf{T.4} & 1 & The benchmark does not modify the environment state. \\
\textbf{T.5} & 1 & Agents have no access to the ground truth information. \\
\textbf{T.6} & 1 & The environment setup is static and does not change over time. \\
\textbf{T.7} & 1 & The data annotation process contains a verification step, as discussed in Section 3.4 of the paper. \\
\textbf{T.8} & 1 & The data annotation process contains a verification step, as discussed in Section 3.4 of the paper. \\
\textbf{T.9} & 0 & The benchmark does not include an Oracle solver that can automatically solve all tasks. \\
\textbf{T.10} & 1 & No vulnerabilities are found in the implementation of the benchmark. \\
\textbf{R.1} & 1 & The benchmark is open-sourced and available on HuggingFace. \\
\textbf{R.2} & 0 & The benchmark does not provide an open-source evaluation harness for users. \\
\textbf{R.3} & 0 & The benchmark does not contain measures to prevent data contamination. \\
\textbf{R.4} & 0 & The report does not discuss plans to consistently update tasks over time. \\
\textbf{R.5} & 1 & Such a relationship is clearly stated in Section 3 of the paper. \\
\textbf{R.6} & 1 & As discussed in Section 3 of the paper, the benchmark is designed to evaluate LLM models. \\
\textbf{R.7} & 1 & Section 5 of the paper discusses the efforts, including comparing evaluation with or without human in the loop. \\
\textbf{R.8} & 1 & {Section 6 discusses the potential impact of unavoidable flaws, such as a wrong reasoning trace resulting in a\\ correct answer.} \\
\textbf{R.9} & 0 & The report does not include quantitative analysis to assess the impact of unavoidable flaws. \\
\textbf{R.10} & 0 & The report does not include any metrics about statistical significance. \\
\textbf{R.11} & 0 & The report does not provide any guidance on interpreting results with eval flaws. \\
\textbf{R.12} & 1 & Human performance is reported in Section 4 of the paper. \\
\textbf{R.13} & 1 & The report includes results of search engine, which can be considered a trivial agent. \\
\bottomrule
\end{longtblr}
}
{\scriptsize
\begin{longtblr}[
    caption = {Assessment Report of OSWorld (\href{https://arxiv.org/pdf/2404.07972}{paper}, \href{https://github.com/xlang-ai/OSWorld}{code})},
    label = {tab:sources} ]{colspec={Q[c]Q[l]Q[l]Q[l]},row{1} = {font=\bfseries}, row{odd[2]} = {bg=gray!25}}%
\toprule
Check & Score & Reason \\
\midrule
\textbf{O.g.1} & 1 & {As discussed in Section 3.2 of the paper, the ground truth is verified to include all states that can be achieved\\ after a successful task completion.} \\
\textbf{O.g.2} & 0 & {The state check only verifies the relevant states for the tasks. Agents can potentially perform extra harmful\\ actions that are not checked by the ground truth.} \\
\textbf{O.g.3} & 1 & {As demonstrated in Section 3.2 of the paper, the ground truth involves complex state changes to a software or\\ website.} \\
\textbf{T.1} & 1 & {No external tools are used in the benchmark. Versions of the environment are clearly specified in the README file\\ of the repository.} \\
\textbf{T.2} & 1 & No external APIs are used in the benchmark. \\
\textbf{T.3} & 1 & No external APIs are used in the benchmark. \\
\textbf{T.4} & 1 & {The benchmark uses virtual machines to run the tasks, which ensures that all residual data or state are cleared\\ between runs.} \\
\textbf{T.5} & 1 & Agents and ground truth are isolated from each other via virtual machines. \\
\textbf{T.6} & 0 & The benchmark checks for HTML selectors (like class names or page titles) on live web pages. \\
\textbf{T.7} & 1 & As discussed in Section 3.2 of the paper, the ground truth is verified for correctness by human experts. \\
\textbf{T.8} & 1 & As discussed in Section 3.2 of the paper, each task is verified to be solvable by human experts. \\
\textbf{T.9} & 0 & The benchmark does not include an Oracle solver that can automatically solve all tasks. \\
\textbf{T.10} & 1 & No vulnerabilities are present in the implementation of the benchmark. \\
\textbf{R.1} & 1 & The benchmark is fully open-sourced, as the code is available on GitHub. \\
\textbf{R.2} & 1 & The benchmark offers an open-source evaluation harness for users. \\
\textbf{R.3} & 0 & The benchmark does not include measures to prevent data contamination. \\
\textbf{R.4} & 0 & The report does not include measures or plans to consistently update tasks over time. \\
\textbf{R.5} & 1 & Such a relationship is clearly stated in Section 2 of the paper. \\
\textbf{R.6} & 1 & As discussed in Section 2 of the paper, the evaluation subject is agent frameworks. \\
\textbf{R.7} & 1 & {As discussed in Section 3.2 of the paper, the benchmark uses additional manual verification steps to prevent,\\ identify, and correct flaws.} \\
\textbf{R.8} & 0 & Safety issues of agents are discussed in Section 7 of the paper. \\
\textbf{R.9} & 0 & No quantitative analysis to assess the impact of unavoidable flaws is included in the report. \\
\textbf{R.10} & 0 & The report does not include metrics about statistical significance. \\
\textbf{R.11} & 0 & The report does not provide guidance on interpreting results with eval flaws. \\
\textbf{R.12} & 1 & Human performance is reported in Section 3.4 of the paper. \\
\textbf{R.13} & 0 & The report does not include results of trivial agents. \\
\bottomrule
\end{longtblr}
}
{\scriptsize
\begin{longtblr}[
    caption = {Assessment Report of KernelBench (\href{https://arxiv.org/pdf/2502.10517}{paper}, \href{https://github.com/ScalingIntelligence/KernelBench/tree/main}{code})},
    label = {tab:sources} ]{colspec={Q[c]Q[l]Q[l]Q[l]},row{1} = {font=\bfseries}, row{odd[2]} = {bg=gray!25}}%
\toprule
Check & Score & Reason \\
\midrule
\textbf{O.e.1} & 0 & The fuzzer does not address potential edge cases, such as empty inputs. \\
\textbf{O.e.2} & 0 & {Although the data type is specified, the fuzzer does not test different memory layouts, such as tensors with\\ non-contiguous memory layouts.} \\
\textbf{O.e.3} & 0 & {The fuzzer uses uniform sampling to generate inputs, which may not be sensitive to the code under testing. For\\ example, the fuzzer may not generate positive inputs that trigger the `relu` function in the `torch` library.} \\
\textbf{T.1} & 0 & The CUDA version is not specified in the default prompt. \\
\textbf{T.2} & 1 & External APIs are not required for the evaluation of the benchmark. \\
\textbf{T.3} & 1 & External APIs are not required for the evaluation of the benchmark. \\
\textbf{T.4} & 1 & Kernels are evaluated in separate processes, and the state is cleared between runs. \\
\textbf{T.5} & 0 & {The ground-truth kernel is executed first and in the same process as the agent. This may lead to the agent\\ accessing the ground-truth results by accessing out-of-bound memory.} \\
\textbf{T.6} & 1 & The environment setup is static and does not change over time. \\
\textbf{T.7} & 1 & The ground-truth kernel is provided by PyTorch, which is a widely used library for deep learning. \\
\textbf{T.8} & 1 & The implementation from PyTorch is a proof of concept. \\
\textbf{T.9} & 1 & The Oracle solver is PyTorch implementation. \\
\textbf{T.10} & 1 & No vulnerabilities are found in the implementation of the benchmark. \\
\textbf{R.1} & 1 & The benchmark is open-sourced and available on GitHub. \\
\textbf{R.2} & 1 & The benchmark provides an open-source evaluation harness for users. \\
\textbf{R.3} & 0 & The benchmark does not discuss measures to prevent data contamination. \\
\textbf{R.4} & 0 & The benchmark does not discuss plans to consistently update tasks over time. \\
\textbf{R.5} & 1 & Section 3 clearly states such a relationship. \\
\textbf{R.6} & 1 & Section 5 clearly states that the evaluation subjective of the benchmark is LLM models. \\
\textbf{R.7} & 1 & {Appendix B.2 describes the efforts taken to prevent, identify, and correct flaws, although these efforts are not\\ sufficient.} \\
\textbf{R.8} & 1 & {Appendix B.2 includes qualitative discussions of the potential impact of unavoidable flaws, although these\\ discussions are not sufficient.} \\
\textbf{R.9} & 1 & {Appendix B.2 includes quantitative analysis to assess the impact of unavoidable flaws, although these analyses are\\ not sufficient.} \\
\textbf{R.10} & 0 & The benchmark does not report any metrics about statistical significance. \\
\textbf{R.11} & 0 & The benchmark does not provide any guidance on interpreting results with eval flaws. \\
\textbf{R.12} & 0 & The benchmark does not report results of non-AI baselines. \\
\textbf{R.13} & 0 & The benchmark does not report results of trivial agents. \\
\bottomrule
\end{longtblr}
}

%% file: tex/app-case_study.tex
\section{Case Study} \label{sec:cases}

We present case study of specific issues we identified. For each study, we use
an Intel E5-2630 CPU with 128 GB RAM and optionally 1 NVIDIA H100 80GB for 
GPU-required experiments. We release our code at \url{https://github.com/uiuc-kang-lab/agentic-benchmarks}.

\subsection{\swebench}

\minihead{Benchmark Overview}
\swebench is a benchmark for evaluating the ability of AI agents to resolve 
real-world GitHub issues. Given the issue description and a summary of the 
codebase, agents are tasked with generating a patch that resolves the issue. 
Each generated patch is evaluated via existing unit tests in the GitHub 
repository.

\minihead{Identified Issue}
SWE-bench uses manually written unit tests to evaluate the correctness of a 
generated code patch. As illustrated in prior work, UTBoost \cite{utboost}, unit 
tests can lead to many false positives, due to the insufficiency of test cases.

\minihead{Example}
The Python package \texttt{seaborn} has an issue in handling missing values in 
the inputs \texttt{x} and \texttt{y} when computing polynomial fits using 
\texttt{PolyFit()}. Unfortunately, the unit test case for \texttt{PolyFit()} 
only considers the scenarios when both \texttt{x} and \texttt{y} have missing 
values:

\begin{lstlisting}[language=Python]
def test_missing_data(self, df):
    groupby = GroupBy([ "group" ])
    df.iloc[5:10] = np.nan
    res1 = PolyFit()( df[[ "x", "y" ]], groupby, "x", {})
    res2 = PolyFit()( df[[ "x" , "y" ]].dropna (), groupby, "x", {})
    assert_frame_equal( res1, res2 )
\end{lstlisting}

This insufficient test case for \texttt{PolyFit()} leads to the following 
incorrect patch for \texttt{PolyFit()} being evaluated as correct. This patch is 
generated by IBM SWE-1.0.

\begin{lstlisting}[language=Python]
def _fit_predict(self, data) :
    y = data ["y"].dropna()
    x = data ["x"].dropna()
    if x.shape[0] != y.shape[0]:
        raise ValueError("x and y must have the same number of non-missing values")
    if x.nunique() <= self.order :
        # TODO warn ?
    xx = yy = []
\end{lstlisting}

\minihead{Qualitative Results}
As reported in prior work \cite{utboost}, agents can pass evaluations without 
addressing the GitHub issues for 5.3\% and 7.7\% of tasks in the Verified and 
Lite partitions, respectively. These tasks lead to 40.9\% and 24.4\% changes in 
the leaderboard for the Verified and Lite partitions, respectively. Furthermore,
these tasks causes 2.3\% and 1.6\% overestimation of agent performance for
the Verified and Lite partitions, respectively.

\subsection{\taubench}

\minihead{Benchmark Overview} 
\taubench is for evaluation AI agents capability to interact with human users 
and follow domain-specific rules \cite{yaotau}. Given a domain-specific policy, the AI agent 
is tasked to interact with human users and answer user queries.

\minihead{Identified Issue} 
\taubench evaluates the agents' actions based on whether the database state is 
correct and optionally whether the agents' responses contain required text. 
Therefore, on tasks that do not change the database state and do not have 
required texts, agents can get positive evaluation results by doing nothing. On 
tasks that do not change the database state and has a trivial required text, 
such as "4", agents can get positive evaluation results by returning random 
responses or all the data.

\minihead{Example}
A task in \taubench requires agent to process a flight cancellation and refund 
request. An AI agent is supposed to check the detail of the booked flight 
ticket for the user in the database and deny the user request if the ticket is 
non-refundable. This task has no required output. Therefore, as long as the 
data state does not change, the agent will obtain a positive evaluation result. 
In this case, an agent that does nothing can also have a positive evaluation
result.

\minihead{Qualitative Results}
A do-nothing agent that returns immediately can achieve a 38\% and 6.0\% 
pass\verb|^|k or pass@k for any k for Airline and Retail partitions, 
respectively. A spamming agent that outputs all the data can achieve a 40\% and
9.6\% pass\verb|^|k or pass@k for any k for Airline and Retail partitions, 
respectively.

\subsection{\birdbench}

\minihead{Benchmark Overview}
\birdbench is for evaluating the capability of agents to write SQL queries 
\cite{li2023can}. Given a query description in natural language, the agent 
needs to translate it into a SQL query.

\minihead{Identified Issue} 
\birdbench evaluates agent by comparing the execution results of the ground truth
query with the generated query. However, due to the ambiguity of the query 
description, there can be multiple correct queries for the same natural language
description.

\minihead{Example}
A task in \birdbench asks the agent to write a SQL query that can answer the 
question: ``What are the name, independence year, and surface area of the 
country with the smallest population?'' There can be two correct SQL queries:
\begin{lstlisting}[language=SQL]
-- Query 1
SELECT Name, SurfaceArea, IndepYear FROM country 
WHERE Population = (SELECT min(Population) FROM country)
-- Query 2
SELECT Name, SurfaceArea, IndepYear FROM country
ORDER BY Population LIMIT 1
\end{lstlisting}
Query 1 outputs all the country with the smallest population, while Query 2 
outputs one of the country with the smallest population. Although the output of 
two queries are different, they both answer the question.

\subsection{\swelancer}

\minihead{Benchmark Overview}
\swelancer is for evaluating the capability of AI agents to independently 
implement features and fix bugs. \cite{miserendino2025swe} Given a task 
description, agent needs to use Python scripts to interact with the file system 
and modify codebase. 

\minihead{Identified Issue}
\swelancer uses end-to-end testing to evaluate the correctness of agents' 
implementation. Although the test cases are stored in a password-protected 
\texttt{.zip} file, reading the directory structure and updating files within 
the \texttt{.zip} file do not require a password. Therefore, an agent can easily 
locate the test cases and replace them with a naive one, such as 
``\texttt{assert 1==1}''. 

\minihead{Qualitative Results}
An agent that overwrites the test cases in the \texttt{.zip} file can achieve a 
100\% resolve rate without completing the software engineering tasks.

\subsection{\webarena}

\minihead{Benchmark Overview}
\webarena is for evaluating the capability of agents to interact with the web 
\cite{zhouwebarena}. Given a user request, the AI agent need either retrieve the 
required information or fill the given data into the web form correctly.

\minihead{Identified Issue} 
\webarena uses exact string matching, substring matching, and LLM-as-a-Judge to 
evaluate agents. Its strategy of exact string matching cannot handle 
alternative expressions and phrase modifiers, while the substring matching is 
vulnerable to exhaustive enumeration of the content on the website. 
Additionally, LLM-as-a-Judge can produce unreliable results.

\minihead{Example}
In \webarena, there is a user query that asks ``What is the duration required to 
first walk from Massachusetts Institute of Technology to Harvard University, and 
then drive to Boston Logan International Airport?'' The ground truth answer for 
this question is 63 minutes. However, the agent searched the web and output the 
final answer: ``The duration required to first walk from Massachusetts Institute 
of Technology to Harvard University is 45 minutes, and then drive to Boston 
Logan International Airport is 8 minutes.'' The answer of agent gives the 
duration of 45+8=53 minutes, which is different from the ground truth answer. 
However, the LLM judge considers the agent's answer as correct.

\subsection{\kernelbench}

\minihead{Benchmark Overview}
\kernelbench is for evaluating the capability of agents to write correct and 
efficient GPU kernels \cite{ouyang2025kernelbench}. Given the task instruction
and the original PyTorch code, agents need to write PyTorch code containing
an inline implementation of the kernel that is functionally correct and more 
efficient.

\minihead{Identified Issue 1}
\kernelbench uses randomly generated inputs (\ie, fuzzing) to test the 
correctness of generated GPU kernels. However, we find the tested functions in a 
subset of tasks are not sensitive to uniform random inputs, such as 
\texttt{mean(softmax(x))} and \texttt{relu(x-2)}. 

\minihead{Identified Issue 2}
In the evaluation implementation, \kernelbench first runs the ground truth 
kernel and then runs the generated kernel subsequently. As reported in prior 
work \cite{lange2025ai}, agents can potentially cheat by generating a program 
that extracts the execution results of the ground truth kernel.

\minihead{Identified Issue 3}
The fuzzer designed in \kernelbench fails to address potential inputs with 
different memory layouts (\eg, non-contiguous tensors), tensor shapes, and 
hardware environment. In the following code snippet, we demonstrate an incorrect
kernel function due to improper use of threads, which were graded as correct in
\kernelbench. In Line 46, the kernel function accesses parallel execution results
in \texttt{s\_sum} with index from \texttt{tid} to \texttt{nthread}. However, 
when \texttt{nthread > normalized\_size}, this will lead to out-of-bound access 
into uninitialized memory. Namely, a thread-safe guard is required here. 

{\scriptsize
\begin{lstlisting}[language=C++]
#include ...

template <typename scalar_t>
__global__ void layernorm_forward_kernel_opt(
    const scalar_t* __restrict__ input,
    const scalar_t* __restrict__ weight,
    const scalar_t* __restrict__ bias,
    const float eps,
    scalar_t* __restrict__ output,
    const int normalized_size) {

  // Each block processes one outer instance.
  int instance_idx = blockIdx.x;

  // Use 2D thread indexing to cover the normalized dimension flexibly.
  int tid = threadIdx.y * blockDim.x + threadIdx.x;
  int nthreads = blockDim.x * blockDim.y;

  // Pointers to the start of this instance's data.
  const scalar_t* __restrict__ in_ptr = input + instance_idx * normalized_size;
  scalar_t* __restrict__ out_ptr = output + instance_idx * normalized_size;

  using accscalar_t = at::acc_type<scalar_t, true>;

  // Each thread computes a partial sum and sum of squares over a strided range.
  accscalar_t local_sum = 0;
  accscalar_t local_sum_sq = 0;
  for (int i = tid; i < normalized_size; i += nthreads) {
    // Use __ldg for read-only, coalesced global memory access
    scalar_t val = __ldg(&in_ptr[i]);
    accscalar_t a_val = static_cast<accscalar_t>(val);
    local_sum += a_val;
    local_sum_sq += a_val * a_val;
  }

  // Allocate shared memory for reduction: first part for partial sums, second for sum of squares.
  extern __shared__ char smem[];
  accscalar_t* s_sum = reinterpret_cast<accscalar_t*>(smem);
  accscalar_t* s_sum_sq = s_sum + nthreads;

  s_sum[tid] = local_sum;
  s_sum_sq[tid] = local_sum_sq;
  __syncthreads();

  // Perform parallel reduction in shared memory.
  for (int stride = nthreads / 2; stride > 0; stride >>= 1) {
    if (tid < stride) {
      s_sum[tid] += s_sum[tid + stride];
      s_sum_sq[tid] += s_sum_sq[tid + stride];
    }
    __syncthreads();
  }
...
}
\end{lstlisting}
}

To identify such issues in large scale, we applied o3-mini to generate additional
test cases. Specifically, we sampled 3 generated kernel functions for each task 
in level 1 and asked o3-mini to detect any possible flaws and write test cases
for each detected flaw. Then, we manually verified the correctness of 
o3-mini-generated test cases. Finally, we applied these test cases on all 
generations by \citet{lange2025ai}. Our results show that the correctness rate
of generated kernels is overestimated by 31\%.

%% file: tex/app-examples.tex
\section{An Example of Rigorous Benchmark Reporting}
\label{sec:app-example}

In this section, we present a modified reporting example based on \birdbench to 
demonstrate benchmark reporting that fulfills all the criteria outlined in 
Figure \ref{fig:report}. \birdbench is a benchmark for evaluating agents' 
capability to translate a natural language query to a SQL query.

\minihead{R.1} Is fully or at least partially open-sourced.

\textbf{Example}: We released the training and validation dataset of \birdbench 
at \url{https://bird-bench.github.io/}.

\minihead{R.2} Offers an open-source evaluation harness for users.

\textbf{Example}: We released the harness to evaluation agents on \birdbench at 
\url{https://github.com/AlibabaResearch/DAMO-ConvAI/tree/main/bird}.

\minihead{R.3} Includes measures to prevent data contamination, such as a
private, held-out test set.

\textbf{Example}: We keep a private held-out test set to avoid potential data 
contamination. Request to evaluate agents on this test set can be submitted at 
\url{https://bird-bench.github.io/}.

\minihead{R.4} Includes measures or plans to consistently update challenges over
time to avoid overfitting.

\textbf{Example}: We plan to consistently update the database and natural 
language queries to reflect the real-world queries and avoid overfitting. Our 
updates will be available at \url{https://bird-bench.github.io/}.

\minihead{R.5} Clearly states the relationship between the agent capabilities it
aims to evaluate and the constructs or outcomes it measures.

\textbf{Example}: \birdbench evaluates agents' capabilities to serve as a 
database interface to translate natural language queries into executable SQL 
queries. To achieve that, \birdbench provides agents with a natural language
query, the database schema, and SQL-related domain knowledge, and challenges 
agents to write a SQL query that can be executed to return correct answers.

\minihead{R.6} Clearly states the evaluation subjective of the benchmark (e.g., a
model or an agent framework).

\textbf{Example}: \birdbench is designed to evaluate the capability of ML models
as well as the performance of agent frameworks.

\minihead{R.7} Describes steps taken to prevent, identify, and correct flaws.

\textbf{Example}: We identify that evaluating generated SQL queries using 
execution results have two limitations. First, tasks requiring \texttt{LIMIT}
queries and containing ties in the data may lead to non-deterministic execution
results. Second, manually annotated ground-truth queries may contain errors. To 
understand and mitigate these errors, we randomly sample 500 tasks to perform
an additional phase of verification. After verifying queries, we found 11.65\% 
of ground-truth queries are incorrect.\footnote[4]{We used results by 
\citet{bird-dev-errors}.}

\minihead{R.8} Includes qualitative discussions of the potential impact of 
unavoidable flaws.

\textbf{Example}: The identified incorrect ground-truth queries and potentially
more incorrect ground-truth queries in the test dataset can lead to estimation
errors of the agent performance and incorrect rankings of agents.

\minihead{R.9} Includes quantitative analysis to assess the impact of unavoidable
flaws (e.g., noise of ground truth).

\textbf{Example}: We build our quantitative analysis based on the normality 
assumption. Specifically, suppose the number of data in the test set $N$ is 
large enough such that the true success rate ($p$) of an agent follows a normal 
distribution with mean $\mu$ and standard deviation $\sigma$. Given the ground 
truth's incorrectness rate of $e$ and the estimated agent success rate $p_0$ 
(based on the imperfect ground truth), $\mu$ and $\sigma$ are calculated as
\begin{equation*}
    \mu = e + (1-2e)p_0; \quad
    \sigma^2 = \mu(1-\mu) = \left(e + (1-2e)p_0\right) \left(1- e - (1-2e)p_0\right)
\end{equation*}
Hence, based on the normality assumption, we can derive a two-sided confidence 
interval with confidence $\alpha$ for $p$ as follows:
\begin{equation}
    \mathbb{P}\left[ \mu - 1.96 \times \frac{\sigma}{\sqrt{N}} \le p \le \mu + 1.96 \times \frac{\sigma}{\sqrt{N}} \right] \ge 95\%
\end{equation}

Finally, based on the plug-in estimate (11.65\%) for the ground truth's 
incorrectness rate, we calculate the confidence interval for the agents' 
performance in Table \ref{tab:bird-data}.

\minihead{R.10} Reports metrics about statistical significance, such as
confidence intervals.

\textbf{Example}: In additional to accuracy estimate, we also calculate 
confidence intervals for each model in Table \ref{tab:bird-data}.

\minihead{R.11} Provides guidance on interpreting results with eval flaws.

\textbf{Example}: Given the potential flaws in \birdbench, we do not recommend 
users to rely on the success rate alone for decision-making or selecting models.
Instead, we suggest using the confidence interval of the success rate as a 
reference.

\minihead{R.12} Reports results of non-AI baselines (e.g., human experts).

\textbf{Example}: We measured the performance of a SQL expert on \birdbench, 
obtaining a success rate of 92.96\%. 

\minihead{R.13} Reports results of trivial agents (e.g., one that does nothing).

\textbf{Example}: We performed sanity check on our evaluation harness by 
measuring the performance of a trivial agent that does nothing. We find that the 
trivial agent achieves 0\% success rate, confirming the rigor of our evaluation
implementation.

\input{figures/tab_bird_bench.tex}

%% file: figures/tab_bird_bench.tex
{\scriptsize
\begin{longtblr}[
    caption = {Modified Leaderboards of \birdbench \cite{li2023can} with Confidence Intervals.},
    label = {tab:bird-data} ]{colspec={Q[c]Q[r]Q[r]Q[r]Q[r]},row{1} = {font=\bfseries}, row{odd[2]} = {bg=gray!25}}%
\toprule
        Method & Dev. Accuracy (\%) & Confidence Interval &  Original Rank & Possible Rank  \\ 
\midrule
CHASE-SQL + Gemini&74.9&[66.8, 71.4]&1&1-13\\
Contextual-SQL&73.5&[65.7, 70.4]&2&1-16\\
XiYan-SQL&73.3&[65.6, 70.2]&3&1-18\\
ExSL + granite-34b-code&72.4&[64.9, 69.6]&4&1-22\\
Reasoning-SQL-14B&72.3&[64.7, 69.4]&5&1-22\\
Insights AI&72.2&[64.6, 69.4]&6&1-22\\
TC-SQL&70.9&[63.7, 68.4]&7&1-27\\
Infly-RL-SQL-32B&70.1&[63.0, 67.8]&8&1-29\\
Queryosity&69.4&[62.5, 67.3]&9&1-32\\
OpenSearch-SQL-v2 + GPT-4o&69.3&[62.4, 67.2]&10&1-32\\
GenaSQL&69.2&[62.4, 67.2]&11&1-33\\
OmniSQL-32B&69.2&[62.4, 67.1]&12&1-33\\
OmniSQL-7B&69.0&[62.2, 67.0]&13&1-33\\
PB-SQL + GPT-4o&68.6&[61.9, 66.7]&14&2-34\\
PURPLE + RED + GPT-4o&68.1&[61.5, 66.3]&15&2-34\\
Arcwise + GPT-4o&68.0&[61.4, 66.2]&16&2-34\\
Distillery + GPT-4o&67.2&[60.8, 65.6]&17&3-36\\
RSL-SQL + GPT-4o&67.2&[60.8, 65.6]&18&3-36\\
XiYanSQL-QwenCoder-32B&67.0&[60.6, 65.5]&19&4-36\\
RECAP + Gemini&67.0&[60.6, 65.4]&20&4-36\\
GSR&66.9&[60.5, 65.4]&21&4-36\\
MSL-SQL + DeepSeek-V2.5&66.8&[60.5, 65.3]&22&4-36\\
AskData + GPT-4o&65.9&[59.8, 64.6]&23&7-37\\
E-SQL + GPT-4o&65.6&[59.5, 64.4]&24&7-37\\
ByteBrain&65.5&[59.4, 64.3]&25&7-37\\
CHESS&65.0&[59.1, 63.9]&26&7-37\\
SCL-SQL&64.7&[58.9, 63.7]&27&7-39\\
EBA-SQL + GPT-4&64.6&[58.8, 63.6]&28&8-39\\
OeSQL-0.1-Qe-32B&64.6&[58.8, 63.6]&29&8-39\\
RSL-SQL + DeepSeek-v2&63.6&[58.0, 62.8]&30&9-42\\
Command-A&63.5&[57.9, 62.8]&31&9-42\\
MCS-SQL + GPT-4&63.4&[57.8, 62.7]&32&9-42\\
PURPLE + GPT-4o&63.0&[57.5, 62.4]&33&11-42\\
GRA-SQL&62.6&[57.2, 62.1]&34&14-44\\
E-SQL + GPT-4o mini&61.6&[56.4, 61.4]&35&17-46\\
OpenSearch-SQL-v1 + GPT-4&61.3&[56.2, 61.2]&36&17-46\\
Dubo-SQL-v1&59.7&[55.0, 59.9]&37&23-49\\
SuperSQL&58.5&[54.0, 59.0]&38&27-49\\
SFT CodeS-15B&58.5&[54.0, 59.0]&39&27-49\\
Chat2Query (GPT-4 + data entity modeling)&58.1&[53.8, 58.7]&40&30-50\\
MAC-SQL + GPT-4&57.6&[53.3, 58.3]&41&30-50\\
SFT CodeS-7B&57.2&[53.0, 58.0]&42&30-51\\
TA-SQL + GPT-4&56.2&[52.3, 57.2]&43&34-51\\
DeepSeek&56.1&[52.2, 57.2]&44&34-51\\
DTS-SQL + DeepSeek-7B&55.8&[52.0, 56.9]&45&35-51\\
SEE&55.5&[51.7, 56.7]&46&35-51\\
DAIL-SQL + GPT-4&54.8&[51.2, 56.1]&47&37-51\\
Interactive-T2S&54.6&[51.0, 56.0]&48&37-51\\
Mistral&53.5&[50.2, 55.2]&49&37-51\\
ExSL + granite-20b-code&51.7&[48.8, 53.8]&50&40-52\\
DIN-SQL + GPT-4&50.7&[48.0, 53.1]&51&42-52\\
GPT-4&46.4&[44.7, 49.7]&52&50-53\\
Claude-2&42.7&[41.9, 46.9]&53&52-54\\
Open-SQL&37.7&[38.1, 43.0]&54&53-54\\
Palm-2&27.4&[30.3, 35.0]&55&55-58\\
ChatGPT + CoT&25.9&[29.2, 33.8]&56&55-58\\
Codex&25.4&[28.8, 33.5]&57&55-58\\
ChatGPT&24.1&[27.8, 32.4]&58&55-58\\
T5-3B&10.4&[17.6, 21.6]&59&59-61\\
T5-Large&9.7&[17.1, 21.1]&60&59-61\\
T5-Base&6.3&[14.6, 18.4]&61&59-61\\
        \bottomrule
    \end{longtblr}
    }

%% file: tex/neurips-checklist.tex

\section{NeurIPS Paper Checklist}

\begin{enumerate}

\item {\bf Claims}
    \item[] Question: Do the main claims made in the abstract and introduction accurately reflect the paper's contributions and scope?
    \item[] Answer: \answerYes{} 
    \item[] Justification: Our claimes in the abstract and introduction are justfied in the later sections and accurately reflect our paper's contribution and scope.

\item {\bf Limitations}
    \item[] Question: Does the paper discuss the limitations of the work performed by the authors?
    \item[] Answer: \answerYes{} 
    \item[] Justification: We included a limitation section in the first section of Appendix (Appendix \ref{sec:app-limit})

\item {\bf Theory Assumptions and Proofs}
    \item[] Question: For each theoretical result, does the paper provide the full set of assumptions and a complete (and correct) proof?
    \item[] Answer: \answerNA{} 
    \item[] Justification: This work does not propose new theories that need proofs.

    \item {\bf Experimental Result Reproducibility}
    \item[] Question: Does the paper fully disclose all the information needed to reproduce the main experimental results of the paper to the extent that it affects the main claims and/or conclusions of the paper (regardless of whether the code and data are provided or not)?
    \item[] Answer: \answerYes{} 
    \item[] Justification: The experiment design and code are open-sourced at \url{https://github.com/uiuc-kang-lab/agentic-benchmarks}. Our experimental results are reproducible with our provided experiment code.

\item {\bf Open access to data and code}
    \item[] Question: Does the paper provide open access to the data and code, with sufficient instructions to faithfully reproduce the main experimental results, as described in supplemental material?
    \item[] Answer: \answerYes{} 
    \item[] Justification: We provide open-sourced code at \url{https://github.com/uiuc-kang-lab/agentic-benchmarks} and open-source data at \url{https://uiuc-kang-lab.github.io/agentic-benchmarks/}. We include detailed justification to our data in Appendix \ref{sec:app-reports}.

\item {\bf Experimental Setting/Details}
    \item[] Question: Does the paper specify all the training and test details (e.g., data splits, hyperparameters, how they were chosen, type of optimizer, etc.) necessary to understand the results?
    \item[] Answer: \answerYes{} 
    \item[] Justification: Experiment parameters and details are discussed in Appendix \ref{sec:cases} and included in our open-source code.

\item {\bf Experiment Statistical Significance}
    \item[] Question: Does the paper report error bars suitably and correctly defined or other appropriate information about the statistical significance of the experiments?
    \item[] Answer: \answerNA{} 
    \item[] Justification: All reported numbers are deterministic. Our experiments contain no stochastic elements, so re-running an experiment yields identical outputs; there is therefore no run-to-run variance on which to base error bars. The evaluation metrics are computed against a single, fixed set of manual annotations (gold standard). For these reasons we do not report error bars or p-values; every number in the experiment section is exact and reproducible.

\item {\bf Experiments Compute Resources}
    \item[] Question: For each experiment, does the paper provide sufficient information on the computer resources (type of compute workers, memory, time of execution) needed to reproduce the experiments?
    \item[] Answer: \answerYes{} 
    \item[] Justification: We specify the computer resources to run our experiments in Appendix \ref{sec:cases}.
    
\item {\bf Code Of Ethics}
    \item[] Question: Does the research conducted in the paper conform, in every respect, with the NeurIPS Code of Ethics \url{https://neurips.cc/public/EthicsGuidelines}?
    \item[] Answer: \answerYes{} 
    \item[] Justification: We confirm that our paper conform, in every respect, with the NeurIPS Code of Ethics.

\item {\bf Broader Impacts}
    \item[] Question: Does the paper discuss both potential positive societal impacts and negative societal impacts of the work performed?
    \item[] Answer: \answerYes{} 
    \item[] Justification: We discuss the potential sociatal impacts of our work in Appendix \ref{sec:app-limit}
    
\item {\bf Safeguards}
    \item[] Question: Does the paper describe safeguards that have been put in place for responsible release of data or models that have a high risk for misuse (e.g., pretrained language models, image generators, or scraped datasets)?
    \item[] Answer: \answerNA{} 
    \item[] Justification: Our work does not release data or model that have a high risks for misuse.

\item {\bf Licenses for existing assets}
    \item[] Question: Are the creators or original owners of assets (e.g., code, data, models), used in the paper, properly credited and are the license and terms of use explicitly mentioned and properly respected?
    \item[] Answer: \answerNA{} 
    \item[] Justification: Our work does not use existing assets.

\item {\bf New Assets}
    \item[] Question: Are new assets introduced in the paper well documented and is the documentation provided alongside the assets?
    \item[] Answer: \answerYes{} 
    \item[] Justification: Our released code are well documented with READMEs.

\item {\bf Crowdsourcing and Research with Human Subjects}
    \item[] Question: For crowdsourcing experiments and research with human subjects, does the paper include the full text of instructions given to participants and screenshots, if applicable, as well as details about compensation (if any)? 
    \item[] Answer: \answerNA{} 
    \item[] Justification: Our paper does not involve crowdsourcing nor human subjects.

\item {\bf Institutional Review Board (IRB) Approvals or Equivalent for Research with Human Subjects}
    \item[] Question: Does the paper describe potential risks incurred by study participants, whether such risks were disclosed to the subjects, and whether Institutional Review Board (IRB) approvals (or an equivalent approval/review based on the requirements of your country or institution) were obtained?
    \item[] Answer: \answerNA{} 
    \item[] Justification: Our paper does not involve crowdsourcing nor human subjects.

\end{enumerate}